\documentclass{article}
\usepackage{algorithm}

\PassOptionsToPackage{numbers,sort&compress}{natbib}
\usepackage[preprint]{neurips_2025}

\usepackage[utf8]{inputenc} 
\usepackage[T1]{fontenc}    
\usepackage{hyperref}       
\usepackage{url}            
\usepackage{booktabs}       
\usepackage{amsfonts}       
\usepackage{nicefrac}       
\usepackage{microtype}      
\usepackage{amsmath}
\usepackage{wrapfig}
\usepackage[utf8]{inputenc}
\usepackage{graphicx}  
\usepackage{amssymb}
\usepackage{caption}
\usepackage{xcolor}
\usepackage{multirow}
\usepackage{algpseudocode}
\usepackage{amssymb}
\usepackage{multirow}
\usepackage{xspace}
\usepackage{bm}
\usepackage{array}
\usepackage{makecell}
\usepackage[table]{xcolor}
\providecommand{\cmark}{\textcolor{green!45!black}{$\checkmark$}}
\usepackage{pifont} 
\definecolor{lightblue}{RGB}{228,242,255}
\definecolor{lightgray}{gray}{0.65}
\definecolor{momblue}{RGB}{47,84,150}
\definecolor{momred}{RGB}{192,57,43}
\definecolor{gt}{HTML}{A9A9B2}
\definecolor{baseline}{HTML}{9D84BC}
\definecolor{memix}{HTML}{E2A19A}

\providecommand{\cmark}{\textcolor{green!45!black}{$\checkmark$}}
\providecommand{\xmark}{\textcolor{red!70!black}{$\times$}}
\providecommand{\gbox}[1]{\begingroup\setlength{\fboxsep}{0.5pt}\colorbox{green!18}{#1}\endgroup}
\providecommand{\toksquares}{\textcolor{gray}{\ensuremath{\blacksquare\!\blacksquare\!\blacksquare}}}

\author{%
	Jiacheng Dong$^{2}$\thanks{Equal contribution.} \quad
	Huan Li$^{1}$\footnotemark[1] \quad
	Sicheng Zhou$^{2}$\footnotemark[1] \quad
	Wenhao Hu$^{2}$ \quad
	Weili Xu$^{2}$ \quad
	Yan Wang$^{1}$\thanks{Corresponding author.} \\
	\\[0.6em]
	$^{1}$Institute for AI Industry Research, Tsinghua University \quad
	$^{2}$Zhejiang University
}

\title{MeMix: Writing Less, Remembering More for Streaming 3D Reconstruction}

\begin{document}
\renewcommand{\thefootnote}{\fnsymbol{footnote}}
\maketitle
\setcounter{footnote}{0}
\renewcommand{\thefootnote}{\arabic{footnote}}

\begin{figure}[H]
\vspace{-1em}
  \centering
\includegraphics[width=0.91\textwidth]{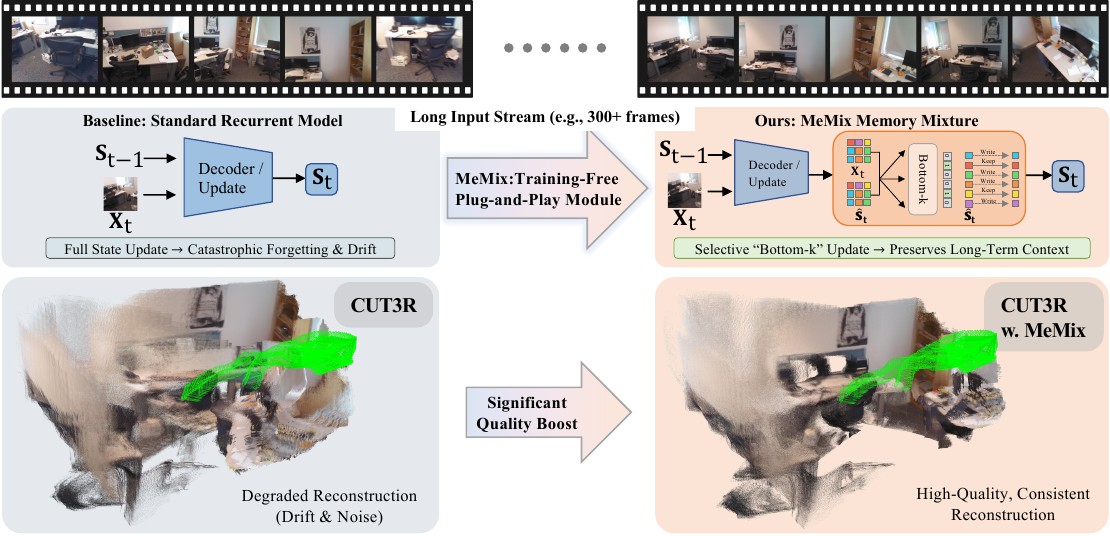}
  \caption{\textbf{MeMix.} A training-free, plug-and-play state-update module for recurrent streaming 3D reconstruction. MeMix recasts the recurrent state as a mixture of memory patches, updates Bottom-k patches and preserves the rest. This reduces interference and improves long-horizon stability with $O(1)$ inference memory.}
  \label{fig:teaser}
\end{figure}

\begin{abstract}
Reconstruction is a fundamental task in 3D vision and a fundamental capability for spatial intelligence. Particularly, streaming 3D reconstruction is central to real-time spatial perception, yet existing recurrent online models often suffer from progressive degradation on long sequences due to state drift and forgetting, motivating inference-time remedies.
We present MeMix, a training-free, plug-and-play module that improves streaming reconstruction by recasting the recurrent state into a \textbf{Me}mory \textbf{Mix}ture.
MeMix partitions the state into multiple independent memory patches and updates only the least-aligned memory patches while exactly preserving others.
This selective update mitigates catastrophic forgetting while retaining $O(1)$ inference memory, and requires no fine-tuning or additional learnable parameters, making it directly applicable to existing recurrent reconstruction models.
Across standard benchmarks (ScanNet, 7-Scenes, KITTI, etc.), under identical backbones and inference settings, MeMix reduces reconstruction completeness error by \textbf{15.3\%} on average (up to \textbf{40.0\%}) across 300--500 frame streams on 7-Scenes. The code is available at \textcolor{momred}{\texttt{https://dongjiacheng06.github.io/MeMix/}}
\end{abstract}
\section{Introduction}
\indent End-to-end 3D reconstruction aims to directly infer camera poses and scene structure from a set of input RGB images, enabling efficient recovery of 3D structure for downstream tasks. Existing methods broadly fall into two paradigms: offline batch reconstruction~\cite{wang2024dust3r,wang2025vggt} and streaming online reconstruction~\cite{wang2025cut3r,wu2025point3r,lan2026stream3r}. Offline batch methods process complete image sequences with global optimization or global-consistency modeling, achieving high reconstruction quality. However, they cannot process arbitrarily long sequences under bounded resources, and their high latency is incompatible with downstream applications that demand real-time spatial perception, such as autonomous driving and robotic navigation~\cite{selvaratnam20253d,liao2025learning}. These constraints motivate streaming online reconstruction, which incrementally consumes a continuously arriving RGB stream and updates geometry and poses in real time.

However, extending online reconstruction to long sequences faces a fundamental tension between exploiting historical context and maintaining constant inference. One family of methods leverages causal-attention KV caches to store the full history~\cite{lan2026stream3r,yuan2026infiniteVGGT,wu2025point3r}. However, this approach incurs memory growth proportional to sequence length, usually leading to out-of-memory errors over long horizons.

An alternative method is fixing latent states to summarize historical context. CUT3R~\cite{wang2025cut3r} formulates reconstruction as a recurrent model with linear attention~\cite{yang2024parallelizing,zhang2026sla2}, achieving $O(1)$ inference memory and computation.  TTT3R~\cite{chen2026ttt3r} reinterprets this under test-time training,  selectively suppressing low-quality updates with adaptive learning rate. Yet since each frame writes into the same set of state tokens, previously stored memories are updated by new information, leading to catastrophic forgetting~\cite{mccloskey1989catastrophic,kirkpatrick2017overcoming}. In practice, this manifests as geometric drift, accumulated pose errors, and degraded long-range consistency.

To address this, we revisit the mixture-of-memories (MoM) idea~\cite{du2025mom} from an engineering perspective. Recent online reconstruction improvements are often presented as standalone methods, and obtaining gains usually requires model-specific redesigns with nontrivial code changes, making reuse across backbones difficult. In contrast, we propose \textbf{MeMix} as a \emph{training-free, plug-and-play} state-update module that can be inserted into existing fixed-state recurrent reconstruction pipelines~\cite{wang2025cut3r,chen2026ttt3r,zheng2026ttsa3r}. Concretely, we partition the state into independent memory patches~\cite{shazeer2017outrageously,fedus2022switch} and update \texttt{Bottom-k} patches at each timestep while the others are preserved. This design reduces cross-time interference without introducing new learnable parameters or any fine-tuning, and we verify clear improvements on three representative baselines.

\textit{Our contribution can be summarized as follows:}
\begin{enumerate}
    \item We introduce MeMix, a training-free, plug-in memory update module that recasts the recurrent state as a mixture of memory patches, substantially improving long-sequence reconstruction quality.
    \item We identify a fundamental bottleneck in fixed-state streaming 3D reconstruction: fully rewriting the recurrent state at each step causes cumulative interference and catastrophic forgetting in long-horizon inference.
    \item MeMix can integrate seamlessly into mainstream recurrent reconstruction models, consistently improving performance with negligible overhead in GPU memory and inference latency.
\end{enumerate}
\section{Related Work}
\noindent \textbf{Feedforward Offline Reconstruction.}
Methods such as DUSt3R~\cite{wang2024dust3r, leroy2024mast3r} encode image features via a ViT~\cite{dosovitskiy2021vit} encoder and achieve 3D matching with cross attention, but they only support pairwise image inputs, and multi-view processing relies on post-processing. VGGT~\cite{wang2025vggt} uses a feedforward Transformer with intra-frame and global self-attention to establish geometric constraints, ensuring global geometric consistency across multiple views. Subsequent methods~\cite{shen2025fastvggt, wang2025flashvggt, sun2025avggt} have further optimized VGGT in inference speed and reconstruction accuracy. However, these offline methods require the full image sequence at inference time and cannot handle arbitrarily long streams under bounded resources, while downstream applications demand real-time perception~\cite{selvaratnam20253d,liao2025learning}. Thus, there is a growing demand for online reconstruction.

\noindent \textbf{Feedforward Online Reconstruction.}
Online reconstruction methods~\cite{wang2025spann3r, wang2025cut3r, wu2025point3r, lan2026stream3r, chen2025long3r} incrementally accept inputs and produce geometry in real time. Existing approaches can be categorized by how they manage historical context. KV-cache based methods~\cite{lan2026stream3r,zhuo2026streamVGGT,mahdi2025evict3r} store historical features in a causal-attention KV cache, retaining long-range context but incurring memory growth with sequence length. Fixed-state methods such as CUT3R~\cite{wang2025cut3r} and TTT3R~\cite{chen2026ttt3r} maintain a fixed-length recurrent state, achieving constant-memory inference. CUT3R interacts input tokens with memory states via cross-attention, but errors accumulate over long sequences due to unconditional full-step writes. TTT3R reinterprets state update as test-time learning~\cite{sun2024learning,mesa}, which eases drift but still suffers from state degradation. Point3R~\cite{wu2025point3r} anchors historical tokens to explicit 3D point positions, achieving strong recall but with memory consumption that grows linearly with the number of views.
While effective, many improvements in this line are tightly coupled to specific architectures and training/inference pipelines, so transferring the same idea to another recurrent backbone often requires substantial re-implementation and engineering effort. This limits practical reusability in real systems where model stacks evolve quickly.

\noindent \textbf{Memory Mixture.}
In sequence modeling~\cite{sutskever2014sequence}, linear attention and state-space models have explored various gating mechanisms to control information retention: RetNet~\cite{sun2023retnet} employs exponential decay for multi-scale retention; Mamba~\cite{gu2023mamba} introduces input-dependent selection for selective state propagation; DeltaNet~\cite{yang2024parallelizing} and Gated DeltaNet~\cite{yang2025gated} adopt the delta rule to address key collisions in additive state updates; and Titans~\cite{behrouz2024titans} introduces a neural long-term memory module that learns to memorize at test time through nested optimization. Subsequent work~\cite{li2025gating,cao2025saga,zhang2024gated} has further deepened the theoretical understanding of how gating controls information flow. However, these approaches all employ continuous gates that never produce exact zeros~\cite{gers2000learning}, meaning every state dimension receives a nonzero update at every step. MoM~\cite{du2025mom} brings sparse routing into linear attention, partitioning the recurrent state into independent memory blocks to reduce cross-time interference.
Our work follows this direction and builds a training-free, plug-and-play state-update module for online 3D reconstruction, validating its effectiveness across three or more recurrent baselines.

\begin{figure}[t]
  \centering
  \includegraphics[width=\textwidth]{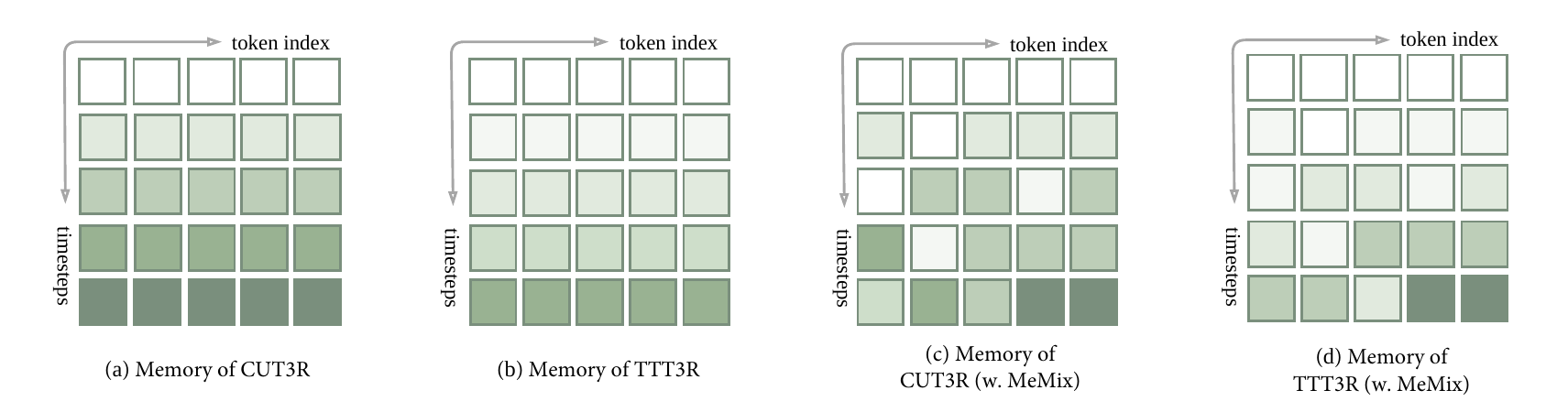}
  \caption{\textbf{Where to write: Mixture Memory Updates.} (a) CUT3R overwrites all state tokens at every timestep. (b) TTT3R applies a dense per-token gate to modulate how much to write, but still updates every token. (c--d) MeMix enables where-to-write updates via Mixture Memory: only a subset of memory patches/tokens are written while the rest are exactly preserved, and it can be plugged into CUT3R (c) or combined with TTT-style gating (d). Colored token squares \toksquares\ indicate tokens that are progressively reinforced over time.}
  \label{fig:visual}
\end{figure}

\section{Method}
MeMix is a training-free method to modify current online 3D reconstruction models~\cite{wang2025cut3r,chen2026ttt3r,zheng2026ttsa3r}. Its core update is executed during the forward pass without any parameter fine-tuning or additional learnable 
modules. Inspired by related works in memory mixture~\cite{du2025mom,behrouz2024titans,liu2024routers,zhang2026sla2}, our key idea is to design the recurrent state as memory blocks~\cite{du2025mom} and selectively update, to reduce cross-time interference under a fixed state.

\subsection{Reconstruction with Continuous Update}
\label{sec:cut3r}

Given a continuous image stream $\{\mathbf{I}_t\}_{t=1}^{T}$, we aim to estimate per-frame camera pose $\mathbf{T}_t$, intrinsic $\mathbf{K}_t$, and pixel-aligned pointmap $\mathbf{P}_t$ in an online fashion. Following CUT3R~\cite{wang2025cut3r}, the process is formulated as a recurrent sequence model operating on a fixed-length state:
\begin{equation}
\begin{aligned}
\mathbf{X}_t = \texttt{Tokenizer}(\mathbf{I}_t) \quad
\mathbf{S}_t = \texttt{Update}(\mathbf{S}_{t-1},\,\mathbf{X}_t) \quad
\mathbf{Y}_t = \texttt{Read}(\mathbf{S}_t,\,\mathbf{X}_t) \quad
\mathcal{M}_t = \texttt{Head}(\mathbf{Y}_t)
\end{aligned}
\label{eq:seq_form}
\end{equation}
where $\mathbf{X}_t\!\in\!\mathbb{R}^{n\times d}$ are image tokens, $\mathbf{S}_t\!\in\!\mathbb{R}^{n\times d}$ is the recurrent state initialized from learnable embeddings, and $\mathbf{Y}_t$ are decoded tokens from which $(\mathbf{T}_t,\mathbf{K}_t,\mathbf{P}_t)$ are regressed.

\subsubsection*{State Input Interaction.}
The \texttt{Update} and \texttt{Read} are realized jointly by an $L$-layer dual-stream cross-attention decoder. Focusing on the state stream, each layer performs:
\begin{equation}
\mathbf{S}^{(\ell)} = \mathbf{S}^{(\ell-1)} + \mathrm{\texttt{softmax}}\!\Big(\mathbf{Q}_{\mathbf{S}}^{(\ell)}\,{\mathbf{K}_{\mathbf{X}}^{(\ell)}}^\top\Big)\,\mathbf{V}_{\mathbf{X}}^{(\ell)}
\label{eq:cross_attn}
\end{equation}
where $\mathbf{Q}_{\mathbf{S}}^{(\ell)}$ is projected from $\mathbf{S}^{(\ell-1)}$ and $\mathbf{K}_{\mathbf{X}}^{(\ell)},\mathbf{V}_{\mathbf{X}}^{(\ell)}$ from $\mathbf{X}^{(\ell-1)}$. A symmetric stream updates $\mathbf{X}^{(\ell)}$ by attending to $\mathbf{S}^{(\ell-1)}$. After $L$ layers, $\mathbf{Y}_t\!=\!\mathbf{X}^{(L)}$ is fed to the prediction head Eq.~\ref{eq:head}.

\subsubsection*{Continuous State Update.}
The continuous update method~\cite{wang2025cut3r} directly takes the decoder output as the new state. Expanding Eq.~\eqref{eq:cross_attn} across $L$ layers:
\begin{equation}
\mathbf{S}_t = \mathbf{S}_{t-1} + \underbrace{\sum_{\ell=1}^{L}\mathrm{\texttt{softmax}}\!\Big(\mathbf{Q}_{\mathbf{S}}^{(\ell)}\,{\mathbf{K}_{\mathbf{X}}^{(\ell)}}^\top\Big)\,\mathbf{V}_{\mathbf{X}}^{(\ell)}}_{\displaystyle\;\Delta\mathbf{S}_t}
\label{eq:cut3r_update}
\end{equation}
where $\Delta\mathbf{S}_t$ is the accumulated cross-attention residual~\cite{zhang2018image,kani2017dr,he2016deep}. This unconditional full-step write, where information from earlier frames is erased by new features, leads to geometric degradation on long sequences.

\subsubsection*{Test-Time Learning for State Update.}
Another method~\cite{chen2026ttt3r} reinterprets the state update through the lens of test-time training. Viewing $\mathbf{S}$ as model parameters and each incoming frame $\mathbf{X}_t$ as a test sample, $\Delta\mathbf{S}_t$ in Eq.~\eqref{eq:cut3r_update} can be seen as a gradient step~\cite{mesa} that minimizes a self-supervised loss on $\mathbf{X}_t$. This method derives $\boldsymbol{\beta}_t$ by aggregating the attention map and scales the entire residual $\Delta\mathbf{S}_t$ before it is added back to $\mathbf{S}_{t-1}$. As shown in Eq.~\eqref{eq:ttt_beta}
\begin{equation}
\boldsymbol{\beta}_t = \sigma\!\Bigg(\frac{1}{LHm}\sum_{\ell,h,j}    Q_{S_{t-1}} \cdot K_{X_t}^T\Bigg)
\label{eq:ttt_beta}
\end{equation}

\begin{figure}[t]
  \centering
  \includegraphics[width=\textwidth]{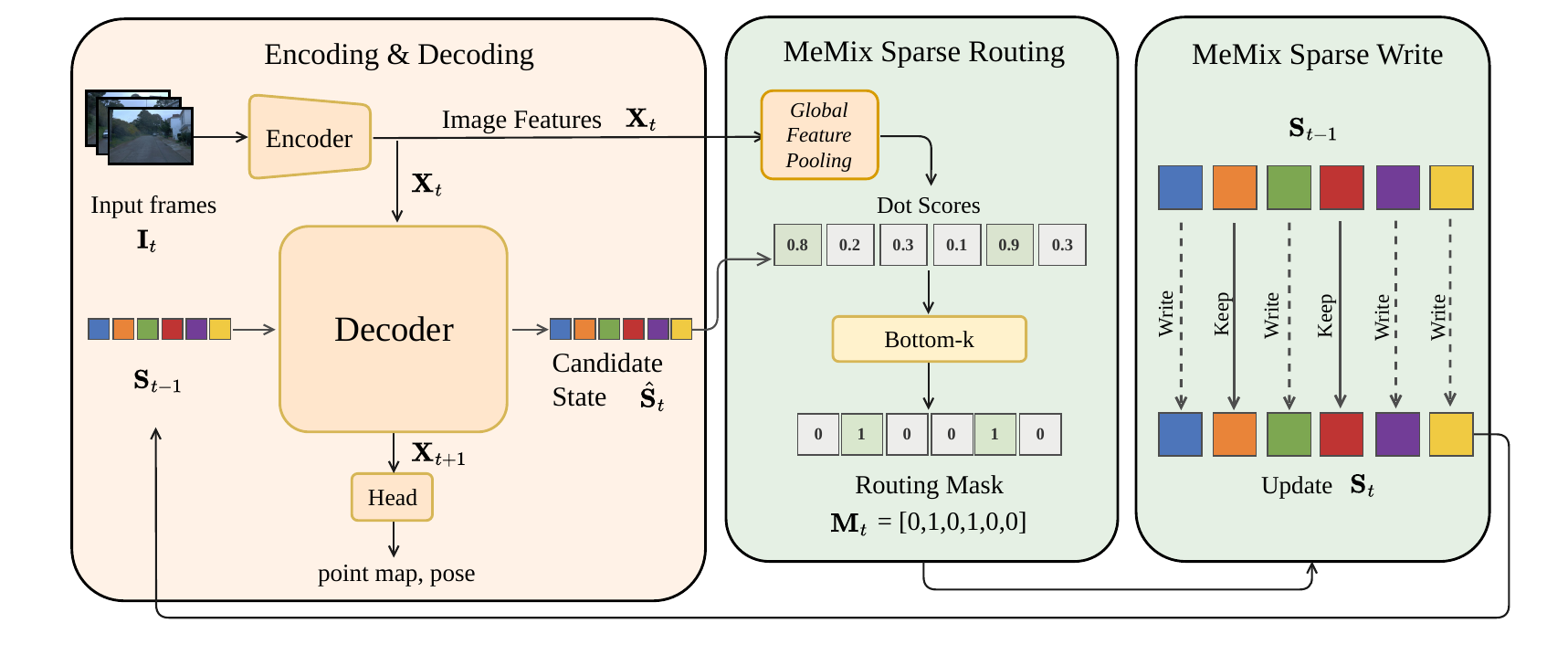}
  \caption{\textbf{Overview of MeMix.} A ViT encoder encodes each frame to tokens $\mathbf{X}_t$, which interact with state tokens $\mathbf{S}_{t-1}$ through a dual-stream cross-attention decoder to produce predictions $\mathbf{Y}_t$ and candidate state $\hat{\mathbf{S}}_t$. MeMix computes dot scores between $\hat{\mathbf{S}}_t$ and ${\mathbf{X}_t}$, selects \texttt{Bottom-k} patches to construct a binary mask $\mathbf{M}_t$, updating only \texttt{Bottom-K} patches. Decoded image tokens $\mathbf{Y}_t$ are fed to the prediction head for output.}
  \label{fig:pipeline}
\end{figure}

\begin{equation}
\mathbf{S}_t 
= \mathbf{S}_{t-1} + \boldsymbol{\beta}_t \odot \sum_{\ell=1}^{L}\mathrm{\texttt{softmax}}\!\Big(\mathbf{Q}_{\mathbf{S}}^{(\ell)}\,{\mathbf{K}_{\mathbf{X}}^{(\ell)}}^\top\Big)\,\mathbf{V}_{\mathbf{X}}^{(\ell)}
\label{eq:ttt_update}
\end{equation}
Each state token now retains history in proportion to its relevance to the current observation, alleviating drift. Nevertheless, the gate remains dense, so every token receives a nonzero write at every step, differing only in magnitude.

\subsection{Rethinking Sparse Update}
\label{sec:sparse_attn}

As mentioned above, when the state is continuously updated, the state at layer $\ell$ evolves as Eq.~\eqref{eq:cross_attn}:

\begin{equation}
\mathbf{S}^{(\ell)} = \mathbf{S}^{(\ell-1)} + \mathbf{A}^{(\ell)}\,\mathbf{V}^{(\ell)},\quad
{\mathbf{A}^{(\ell)}=\mathrm{\texttt{softmax}}\!\Big(\mathbf{Q}_{\mathbf{S}}^{(\ell)}\,\notag{\mathbf{K}_{\mathbf{X}}^{(\ell)}}^\top\Big)} 
\end{equation}

After $L$ layers, the decoder produces a candidate state $\hat{\mathbf{S}}_t\!=\!\mathbf{S}^{(L)}$. If there is no gating, the state is directly overwritten: $\mathbf{S}_t\!=\!\hat{\mathbf{S}}_t$ as Eq.~\eqref{eq:cut3r_update}.

Now introduce a gate matrix ${\boldsymbol{{G_t}}\!\in\![0,1]}$ that modulates the attention:

\begin{equation}
\mathbf{S}^{(\ell)} = \mathbf{S}^{(\ell-1)} + (\boldsymbol{G}_t \odot \mathbf{A}^{(\ell)})\,\mathbf{V}^{(\ell)}
\label{eq:gated_attn}
\end{equation}

\begin{wraptable}{r}{0.5\columnwidth}
\vspace{-0.1em}
\centering
\footnotesize
\caption{\textbf{Unified memory update rules.}
The methods in Section~3 can be expressed under a shared gated state update framework.
CUT3R corresponds to full-state overwrite, TTT3R/TTSA3R to dense token-wise gating, MeMix to sparse binary routing, and their combination to sparse routed soft gating.}
\label{tab:memory_update}
\renewcommand{\arraystretch}{1.1}
\setlength{\tabcolsep}{3pt}
\begin{tabular}{p{0.27\columnwidth} p{0.20\columnwidth}}
\toprule
\textbf{Method} & \textbf{Rule} \\
\midrule
Unified & \hspace{-0.8em}\makecell[l]{$S_t = G_t \odot \hat S_t$ \\
$ + (1-G_t)\odot S_{t-1}$} \\
\midrule
CUT3R & $G_t = 1$ \\
TTT3R/TTSA3R & $G_t = \beta_t$ \\
CUT3R + MeMix & $G_t = M_t$ \\
TTT3R/TTSA3R + MeMix & $G_t = M_t \odot \beta_t$ \\
\bottomrule
\end{tabular}
\vspace{-0.6em}
\end{wraptable}

Substituting into the residual connection, the gated candidate state becomes

\begin{equation}
\hat{\mathbf{S}}_{t}\!=\!\mathbf{S}_{t-1}+\sum_{\ell=1}^{L} (\boldsymbol{G}_t \odot\mathbf{A}^{(\ell)})\mathbf{V}^{(\ell)}
\label{eq:sum_gated_attn}
\end{equation}

The state update can then be written as

\begin{equation}
\mathbf{S}_t = \boldsymbol{G}_t \odot \hat{\mathbf{S}}_t + (1-\boldsymbol{G}_t)\odot \mathbf{S}_{t-1}
\label{eq:gate_update1}
\end{equation}

\begin{equation}
\mathbf{S}_t = \mathbf{S}_{t-1}+\boldsymbol{G}_t \sum_{\ell=1}^{L}\mathbf{A}^{(\ell)}\,\mathbf{V}^{(\ell)}
\label{eq:gate_update2}
\end{equation}

when $\boldsymbol{G}_t\!\in\!(0,1)$, Eq.~\eqref{eq:gate_update2} is equivalent to Eq.~\eqref{eq:ttt_update}

when $\boldsymbol{G}_t\!\in\![0,1]$, we have the sparse update rule.

Table~\ref{tab:memory_update} summarizes how CUT3R, TTT3R/TTSA3R, and MeMix-based variants can all be written under a shared gate formulation with different instantiations of $\boldsymbol{G}_t$.
\subsection{MeMix Design}
\label{sec:MeMix}

\subsubsection*{Memory Mixture}
Unlike the full update strategy of CUT3R and the dense learning-rate adaptation of TTT3R, MeMix constructs a routing mask $\mathbf{M}_t$~\cite{zhang2026sla2} for the state update in Eq.~\eqref{eq:gate_update1}. In step $t$, the decoder produces a candidate state $\hat{\mathbf{S}}_t\!\in\!\mathbb{R}^{n\times d}$ and image features $\mathbf{X}_t\!\in\!\mathbb{R}^{n\times d}$. The interaction between state token and the observation is measured by dot-product similarity:

\begin{equation}
r_{t} = \langle\,\hat{\mathbf{S}}_{t},\;{\mathbf{X}}_t\rangle
\label{eq:token_score}
\end{equation}

Then, we select the k patches with the bottom scores:
\begin{equation}
\mathrm{\mathcal{P}_t = \texttt{Bottom}\text{-}}\texttt{k}\!\big\{r_{t}\}
\label{eq:routing}
\end{equation}

The routing mask is then constructed from the selected Bottom-$k$ patches, where
$M_t=1$ marks the subset to be updated. Therefore, the updated subset
corresponds to the least-aligned (Bottom-$k$) patches according to the routing
score $r_t$, while the remaining higher-score patches are preserved.

\noindent
\begin{minipage}[t]{0.52\textwidth}
\vspace{0pt}
\subsubsection*{State Update}
Substituting the routing mask $\mathbf{M}_t$ into Eq.~\eqref{eq:gate_update1}:
\begin{equation}
\mathbf{S}_t = \mathbf{M}_t \odot \hat{\mathbf{S}}_t + (1-\mathbf{M}_t)\odot \mathbf{S}_{t-1}
\label{eq:mom_update}
\end{equation}
Tokens within the selected patches are fully replaced by the decoder output; all others are exactly preserved. This is the binary-gate instance of Eq.~\eqref{eq:gate_update1} discussed in Sec.~\ref{sec:sparse_attn}.

\subsubsection*{State Update with Test-Time Training} Routing mask $\mathbf{M}_t$ determines where to write, yet within the selected patches every token is still fully overwritten. We can further modulate how much to write by combining $\mathbf{M}_t$ with the attention-derived learning rate $\boldsymbol{\beta}_t$ from Eq.~\eqref{eq:ttt_beta}:
\begin{equation}
\mathbf{S}_t = (\mathbf{M}_t \odot \boldsymbol{\beta}_t) \odot \hat{\mathbf{S}}_t + (1 - \mathbf{M}_t \odot \boldsymbol{\beta}_t)\odot \mathbf{S}_{t-1}
\label{eq:mom_beta_update}
\end{equation}
\end{minipage}
\hfill
\begin{minipage}[t]{0.45\textwidth}
\vspace{0pt}
\begin{tabular}{@{}p{\linewidth}@{}}
\toprule
\textbf{Algorithm. MeMix Inference} \\
\midrule
\begin{algorithmic}[1]
\Require Input sequence $\{I_t\}_{t=1}^T$, base model $f$, patch partition $\{P_j\}_{j=1}^p$
\State \textcolor{momblue}{$S \gets S_0$}
\State \textcolor{momblue}{$\mathcal{M} \gets \emptyset$}
\For{\textcolor{momblue}{$t = 1$ to $T$}}
    \State \textcolor{momblue}{$X_t \gets \text{\texttt{Tokenize}}(I_t)$}
    \State \textcolor{momblue}{$\hat S_{t}, Y_{t} \gets \text{cross attn}( S_{t-1},{X}_{t})$}
    \State \textcolor{momred}{$r_{t} \gets \text{RouteScore}( \hat S_{t},{X}_t)$}
    \State \textcolor{momred}{$\mathcal{P}_t \gets \text{\texttt{Bottom-K}}(r_{t})$}
    \State \textcolor{momred}{$\mathbf{M_t} \gets \text{Gate}(\mathcal{P}_t)$}
    \State \textcolor{momred}{$S_t \gets \mathbf{M_t} \odot \hat{S_{t}} + (1-\mathbf{M_t}) \odot S_{t-1}$}
    \State \textcolor{momblue}{$\mathcal{M}_t \gets {\texttt{Head}(Y_t)}$}
\EndFor
\State \textcolor{momblue}{\textbf{return} $\mathcal{M}$}
\end{algorithmic} \\
\bottomrule
\end{tabular}
\end{minipage}

For unselected patches $M_{t}\!=\!0$, the state is exactly preserved regardless of $\boldsymbol{\beta}_t$; for selected patches, $\boldsymbol{\beta}_t$ provides token-level soft modulation. This combination is still training-free since $\boldsymbol{\beta}_t$ is derived from the existing decoder's cross-attention weights. Note that $\mathbf{M}_t \odot \boldsymbol{\beta}_t \in [0,1]$, so Eq.~\eqref{eq:mom_beta_update} remains equivalent to the unified gate framework in Eq.~\eqref{eq:gate_update1}.

\subsubsection*{Readout and Output}Symmetrically to the state stream Eq.~\eqref{eq:cross_attn}, the image stream of the dual-stream decoder lets each image token cross-attend to the state at every layer:

\begin{equation}
\mathbf{X}^{(\ell)} = \mathbf{X}^{(\ell-1)} + \texttt{softmax}\!\Big(\mathbf{Q}_{\mathbf{X}}^{(\ell)}\,{\mathbf{K}_{\mathbf{S}}^{(\ell)}}^{\!\top}\Big)\,\mathbf{V}_{\mathbf{S}}^{(\ell)}
\end{equation}
\begin{equation}
\mathbf{X} = \mathbf{X}_{t-1} + \sum_{\ell=1}^{L}\texttt{softmax}\!\Big(\mathbf{Q}_{\mathbf{X}}^{(\ell)}\,{\mathbf{K}_{\mathbf{S}}^{(\ell)}}^{\!\top}\Big)\,\mathbf{V}_{\mathbf{S}}^{(\ell)}
\label{eq:readout}
\end{equation}
where $\mathbf{Q}_{\mathbf{X}}^{(\ell)}$ is projected from $\mathbf{X}^{(\ell-1)}$ and $\mathbf{K}_{\mathbf{S}}^{(\ell)},\mathbf{V}_{\mathbf{S}}^{(\ell)}$ from $\mathbf{S}^{(\ell-1)}$.
After $L$ layers, the decoded tokens $\mathbf{Y}_t\!=\!\mathbf{X}$ are fed to the DPT head~\cite{ranftl2021vision,chen2022vision}. Finally we have:
\begin{equation}
(\mathbf{T}_t,\,\mathbf{K}_t,\,\mathbf{P}_t) = \texttt{Head}(\mathbf{Y}_t)
\label{eq:head}
\end{equation}

\section{Experiments}
\label{sec:exp}

We evaluate MeMix on multi-view 3D reconstruction, camera pose estimation, and video depth estimation. As a \emph{training-free} and \emph{plug-and-play} module, we keep the backbone weights, input resolution, and inference hyper-parameters identical to the corresponding baseline.

\textbf{Baselines.}
Following common practice in recent streaming reconstruction evaluations \cite{wang2025cut3r,chen2026ttt3r,zheng2026ttsa3r}, we compare MeMix with representative offline and online families. We first include strong pairwise 3D reconstruction foundation models, including DUSt3R \cite{wang2024dust3r}, MASt3R \cite{leroy2024mast3r}, MonST3R \cite{zhang2025monst3r}, and Easi3R \cite{chen2025easi3r}, which take a pair of views as input and typically require an extra global alignment stage to consolidate pairwise predictions. We also compare with full-attention multiview models such as AETHER \cite{zhu2025aether} and VGGT \cite{wang2025vggt}, which can jointly predict pointmaps/cameras but are usually limited to long sequences due to the need to run full attention when new frames arrive. 

For online methods, we plug MeMix into recurrent streaming backbones and compare the results with their original versions, including CUT3R \cite{wang2025cut3r} and its training-free adaptations TTT3R \cite{chen2026ttt3r} and TTSA3R \cite{zheng2026ttsa3r}. We further include KV-cache / causal streaming Transformers (StreamVGGT \cite{zhuo2026streamVGGT} and STREAM3R$^\alpha$ \cite{lan2026stream3r}) to test our overall competitiveness. Finally, we compare with explicit/external memory designs (Spann3R \cite{wang2025spann3r}, Point3R \cite{wu2025point3r}), which improve long-horizon recall by maintaining additional pointmap memories.

\textbf{Settings.}
In all experiments, we run inference at a single NVIDIA A100 40GB PCIe or RTX 4090 (24GB). The \texttt{Bottom-k} is set to 708 while the entire state is 768 tokens/frame---this is the most fine-grained design, ensuring that every token can be involved. Following prior training-free methods\cite{chen2026ttt3r,zheng2026ttsa3r,shen2025mut3r}, we apply MeMix to existing pipelines, using the same released checkpoint and evaluation script, without any fine-tuning. Computation costs are shown in Table~\ref{table:5}.






\subsection{3D Reconstruction}
\label{sec:exp_recon}

Following common practice in long-horizon streaming reconstruction \cite{wang2025cut3r,chen2026ttt3r,zhuo2026streamVGGT,yuan2026infiniteVGGT}, we evaluate multi-view reconstruction on 7-Scenes \cite{dataset7scenes} and NRGBD \cite{datasetnrgbd}. To explicitly probe length generalization under bounded memory, we tested three long sequence lengths (300/400/500 frames) and reported accuracy, completeness, and normal consistency.

As shown in Table~\ref{tab:mv_recon_long}, offline full-attention models (e.g. VGGT) and KV-cache streaming Transformers (e.g. StreamVGGT) run out of memory at all tested lengths (300/400/500 frames). For constant-memory recurrent baselines, reconstruction quality generally degrades as the input horizon increases. In contrast, inserting MeMix into the same backbone consistently improves 7-Scenes performance in accuracy, completeness, and normal consistency for all lengths tested. In NRGBD, MeMix also provides overall gains across all backbones and input lengths, with especially clear improvements in accuracy and completeness.

\begin{figure}[t]
  \centering
  \includegraphics[width=\textwidth]{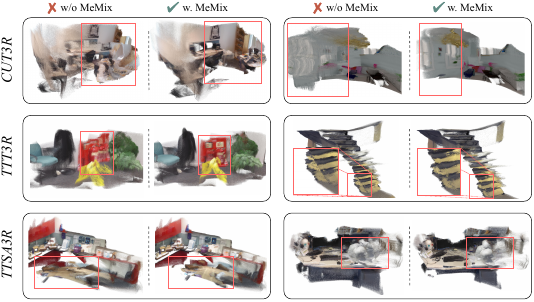}
  \caption{\textbf{Qualitative results of 3D reconstruction.} We compare CUT3R, TTT3R, and TTSA3R with their MeMix variants on long input streams. MeMix consistently improves reconstruction quality by reducing drift and recovering more complete, sharper surfaces. Red boxes highlight representative regions where MeMix corrects failures such as surface tearing, missing geometry, and ghosting.}
  \label{fig:demos}
\end{figure}

Fig.~\ref{fig:demos} shows comparisons between methods with or without MeMix. Without MeMix, recurrent baselines accumulate errors over time, which typically appear as pose drift and degraded geometry. MeMix leads to more coherent surfaces and better preserved structures. More results are shown in Table ~\ref{S-tab:mv_recon_long} of the supplementary material.

\begin{table}[h]
\centering
\caption{
    \textbf{3D Reconstruction Results on 7-Scenes~\cite{dataset7scenes} and NRGBD~\cite{datasetnrgbd}.}
     We test MeMix on 7-Scenes and NRGBD, with one frame sampled every two frames (Sparse Sampling, \textbf{-S}). \protect\gbox{Green boxes} indicate improved or unchanged performance over the base model (w/o MeMix) under the same input length.
}
\resizebox{\textwidth}{!}{
\begin{tabular}{lcccccccccccccc}
    \toprule[0.17em]
    {\multirow{4}{*}{\textbf{Model}}} &
    {\multirow{4}{*}{\textbf{MeMix}}} &
    {\multirow{4}{*}{\textbf{Input}}} &
    \multicolumn{6}{c}{\textbf{7-Scenes-S}} &
    \multicolumn{6}{c}{\textbf{NRGBD-S}} \\
    \cmidrule(r){4-9} \cmidrule(r){10-15}
    & & &
    \multicolumn{2}{c}{Acc. $\downarrow$} &
    \multicolumn{2}{c}{Comp. $\downarrow$} &
    \multicolumn{2}{c}{NC $\uparrow$} &
    \multicolumn{2}{c}{Acc. $\downarrow$} &
    \multicolumn{2}{c}{Comp. $\downarrow$} &
    \multicolumn{2}{c}{NC $\uparrow$} \\
    \cmidrule(r){4-5} \cmidrule(r){6-7} \cmidrule(r){8-9}
    \cmidrule(r){10-11} \cmidrule(r){12-13} \cmidrule(r){14-15}
    & & &
    Mean & Med. &
    Mean & Med. &
    Mean & Med. &
    Mean & Med. &
    Mean & Med. &
    Mean & Med. \\
    \midrule[0.08em]

\multirow{3}{*}{VGGT \textit{(Offline)}~\cite{wang2025vggt}} & -- & \textit{300} &
\textcolor{lightgray}{\textit{OOM}} & \textcolor{lightgray}{\textit{OOM}} &
\textcolor{lightgray}{\textit{OOM}} & \textcolor{lightgray}{\textit{OOM}} &
\textcolor{lightgray}{\textit{OOM}} & \textcolor{lightgray}{\textit{OOM}} &
\textcolor{lightgray}{\textit{OOM}} & \textcolor{lightgray}{\textit{OOM}} &
\textcolor{lightgray}{\textit{OOM}} & \textcolor{lightgray}{\textit{OOM}} &
\textcolor{lightgray}{\textit{OOM}} & \textcolor{lightgray}{\textit{OOM}} \\
& -- & \textit{400} &
\textcolor{lightgray}{\textit{OOM}} & \textcolor{lightgray}{\textit{OOM}} &
\textcolor{lightgray}{\textit{OOM}} & \textcolor{lightgray}{\textit{OOM}} &
\textcolor{lightgray}{\textit{OOM}} & \textcolor{lightgray}{\textit{OOM}} &
\textcolor{lightgray}{\textit{OOM}} & \textcolor{lightgray}{\textit{OOM}} &
\textcolor{lightgray}{\textit{OOM}} & \textcolor{lightgray}{\textit{OOM}} &
\textcolor{lightgray}{\textit{OOM}} & \textcolor{lightgray}{\textit{OOM}} \\
& -- & \textit{500} &
\textcolor{lightgray}{\textit{OOM}} & \textcolor{lightgray}{\textit{OOM}} &
\textcolor{lightgray}{\textit{OOM}} & \textcolor{lightgray}{\textit{OOM}} &
\textcolor{lightgray}{\textit{OOM}} & \textcolor{lightgray}{\textit{OOM}} &
\textcolor{lightgray}{\textit{OOM}} & \textcolor{lightgray}{\textit{OOM}} &
\textcolor{lightgray}{\textit{OOM}} & \textcolor{lightgray}{\textit{OOM}} &
\textcolor{lightgray}{\textit{OOM}} & \textcolor{lightgray}{\textit{OOM}} \\
\midrule[0.08em]

\multirow{3}{*}{StreamVGGT~\cite{zhuo2026streamVGGT}} & -- & \textit{300} &
\textcolor{lightgray}{\textit{OOM}} & \textcolor{lightgray}{\textit{OOM}} &
\textcolor{lightgray}{\textit{OOM}} & \textcolor{lightgray}{\textit{OOM}} &
\textcolor{lightgray}{\textit{OOM}} & \textcolor{lightgray}{\textit{OOM}} &
\textcolor{lightgray}{\textit{OOM}} & \textcolor{lightgray}{\textit{OOM}} &
\textcolor{lightgray}{\textit{OOM}} & \textcolor{lightgray}{\textit{OOM}} &
\textcolor{lightgray}{\textit{OOM}} & \textcolor{lightgray}{\textit{OOM}} \\
& -- & \textit{400} &
\textcolor{lightgray}{\textit{OOM}} & \textcolor{lightgray}{\textit{OOM}} &
\textcolor{lightgray}{\textit{OOM}} & \textcolor{lightgray}{\textit{OOM}} &
\textcolor{lightgray}{\textit{OOM}} & \textcolor{lightgray}{\textit{OOM}} &
\textcolor{lightgray}{\textit{OOM}} & \textcolor{lightgray}{\textit{OOM}} &
\textcolor{lightgray}{\textit{OOM}} & \textcolor{lightgray}{\textit{OOM}} &
\textcolor{lightgray}{\textit{OOM}} & \textcolor{lightgray}{\textit{OOM}} \\
& -- & \textit{500} &
\textcolor{lightgray}{\textit{OOM}} & \textcolor{lightgray}{\textit{OOM}} &
\textcolor{lightgray}{\textit{OOM}} & \textcolor{lightgray}{\textit{OOM}} &
\textcolor{lightgray}{\textit{OOM}} & \textcolor{lightgray}{\textit{OOM}} &
\textcolor{lightgray}{\textit{OOM}} & \textcolor{lightgray}{\textit{OOM}} &
\textcolor{lightgray}{\textit{OOM}} & \textcolor{lightgray}{\textit{OOM}} &
\textcolor{lightgray}{\textit{OOM}} & \textcolor{lightgray}{\textit{OOM}} \\
\midrule[0.08em]

\multirow{6}{*}{CUT3R~\cite{wang2025cut3r}} & \xmark & \textit{300} &
0.141 & 0.096 & 0.076 & 0.034 & 0.543 & 0.564 &
0.234 & 0.139 & 0.074 & 0.018 & 0.575 & 0.614 \\
& \cellcolor{lightblue}\cmark & \cellcolor{lightblue}{} &
\cellcolor{lightblue}\gbox{0.106} & \cellcolor{lightblue}\gbox{0.076} &
\cellcolor{lightblue}\gbox{0.053} & \cellcolor{lightblue}\gbox{0.019} &
\cellcolor{lightblue}\gbox{0.550} & \cellcolor{lightblue}\gbox{0.575} &
\cellcolor{lightblue}\gbox{0.186} & \cellcolor{lightblue}\gbox{0.086} &
\cellcolor{lightblue}\gbox{0.050} & \cellcolor{lightblue}\gbox{0.009} &
\cellcolor{lightblue}\gbox{0.595} & \cellcolor{lightblue}\gbox{0.651} \\
& \xmark & \textit{400} &
0.178 & 0.121 & 0.115 & 0.069 & 0.532 & 0.546 &
0.342 & 0.227 & 0.127 & 0.067 & 0.561 & 0.591 \\
& \cellcolor{lightblue}\cmark & \cellcolor{lightblue}{} &
\cellcolor{lightblue}\gbox{0.147} & \cellcolor{lightblue}\gbox{0.100} &
\cellcolor{lightblue}\gbox{0.076} & \cellcolor{lightblue}\gbox{0.039} &
\cellcolor{lightblue}\gbox{0.540} & \cellcolor{lightblue}\gbox{0.559} &
\cellcolor{lightblue}\gbox{0.321} & \cellcolor{lightblue}\gbox{0.180} &
\cellcolor{lightblue}\gbox{0.099} & \cellcolor{lightblue}\gbox{0.031} &
\cellcolor{lightblue}0.565 & \cellcolor{lightblue}0.594 \\
& \xmark & \textit{500} &
0.190 & 0.138 & 0.090 & 0.033 & 0.530 & 0.543 &
0.359 & 0.264 & 0.173 & 0.081 & 0.560 & 0.591 \\
& \cellcolor{lightblue}\cmark & \cellcolor{lightblue}{} &
\cellcolor{lightblue}\gbox{0.167} & \cellcolor{lightblue}\gbox{0.119} &
\cellcolor{lightblue}\gbox{0.077} & \cellcolor{lightblue}\gbox{0.026} &
\cellcolor{lightblue}\gbox{0.533} & \cellcolor{lightblue}\gbox{0.547} &
\cellcolor{lightblue}\gbox{0.328} & \cellcolor{lightblue}\gbox{0.218} &
\cellcolor{lightblue}\gbox{0.161} & \cellcolor{lightblue}\gbox{0.040} &
\cellcolor{lightblue}\gbox{0.560} & \cellcolor{lightblue}0.590 \\
\midrule[0.08em]

\multirow{6}{*}{TTT3R~\cite{chen2026ttt3r}} & \xmark & \textit{300} &
0.040 & 0.025 & 0.024 & 0.005 & 0.567 & 0.602 &
0.101 & 0.044 & 0.025 & 0.005 & 0.610 & 0.678 \\
& \cellcolor{lightblue}\cmark & \cellcolor{lightblue}{} &
\cellcolor{lightblue}\gbox{0.034} & \cellcolor{lightblue}\gbox{0.020} &
\cellcolor{lightblue}\gbox{0.023} & \cellcolor{lightblue}\gbox{0.005} &
\cellcolor{lightblue}\gbox{0.567} & \cellcolor{lightblue}\gbox{0.603} &
\cellcolor{lightblue}\gbox{0.099} & \cellcolor{lightblue}\gbox{0.037} &
\cellcolor{lightblue}\gbox{0.020} & \cellcolor{lightblue}\gbox{\bf 0.004} &
\cellcolor{lightblue}\gbox{0.616} & \cellcolor{lightblue}\gbox{0.692} \\
& \xmark & \textit{400} &
0.052 & 0.031 & 0.027 & 0.005 & 0.558 & 0.588 &
0.143 & 0.065 & 0.071 & 0.012 & 0.600 & 0.658 \\
& \cellcolor{lightblue}\cmark & \cellcolor{lightblue}{} &
\cellcolor{lightblue}\gbox{0.043} & \cellcolor{lightblue}\gbox{0.025} &
\cellcolor{lightblue}\gbox{0.026} & \cellcolor{lightblue}\gbox{0.005} &
\cellcolor{lightblue}\gbox{0.560} & \cellcolor{lightblue}\gbox{0.590} &
\cellcolor{lightblue}0.146 & \cellcolor{lightblue}0.066 &
\cellcolor{lightblue}\gbox{0.070} & \cellcolor{lightblue} 0.018 &
\cellcolor{lightblue}\gbox{0.602} & \cellcolor{lightblue}\gbox{0.665} \\
& \xmark & \textit{500} &
0.066 & 0.039 & 0.031 & 0.006 & 0.551 & 0.577 &
0.166 & 0.092 & 0.087 & 0.021 & 0.593 & 0.647 \\
& \cellcolor{lightblue}\cmark & \cellcolor{lightblue}{} &
\cellcolor{lightblue}\gbox{0.059} & \cellcolor{lightblue}\gbox{0.032} &
\cellcolor{lightblue}\gbox{0.030} & \cellcolor{lightblue}\gbox{0.005} &
\cellcolor{lightblue}\gbox{0.553} & \cellcolor{lightblue}\gbox{0.580} &
\cellcolor{lightblue}0.183 & \cellcolor{lightblue}0.094 &
\cellcolor{lightblue}0.094 & \cellcolor{lightblue}0.031 &
\cellcolor{lightblue}\gbox{0.595} & \cellcolor{lightblue}\gbox{0.650} \\
\midrule[0.08em]

\multirow{6}{*}{TTSA3R~\cite{zheng2026ttsa3r}} & \xmark & \textit{300} &
0.036 & 0.020 & 0.035 & 0.006 & 0.566 & 0.600 &
0.090 & 0.036 & 0.020 & \bf 0.004 & 0.620 & 0.696 \\
& \cellcolor{lightblue}\cmark & \cellcolor{lightblue}{} &
\cellcolor{lightblue}\gbox{\bf 0.026} & \cellcolor{lightblue}\gbox{\bf 0.013} &
\cellcolor{lightblue}\gbox{\bf 0.021} & \cellcolor{lightblue}\gbox{\bf 0.004} &
\cellcolor{lightblue}\gbox{\bf 0.568} & \cellcolor{lightblue}\gbox{\bf 0.604} &
\cellcolor{lightblue}\gbox{\bf 0.086} & \cellcolor{lightblue}\gbox{\bf 0.031} &
\cellcolor{lightblue}\gbox{\bf 0.015} & \cellcolor{lightblue}\gbox{\bf 0.004} &
\cellcolor{lightblue}\gbox{\bf 0.626} & \cellcolor{lightblue}\gbox{\bf 0.709} \\
& \xmark & \textit{400} &
0.036 & 0.019 & 0.024 & \bf 0.004 & \bf 0.561 & 0.592 &
0.104 & 0.045 & 0.035 & 0.006 & \bf 0.618 & \bf 0.692 \\
& \cellcolor{lightblue}\cmark & \cellcolor{lightblue}{} &
\cellcolor{lightblue}\gbox{\bf 0.030} & \cellcolor{lightblue}\gbox{\bf 0.015} &
\cellcolor{lightblue}\gbox{\bf 0.023} & \cellcolor{lightblue}\gbox{\bf 0.004} &
\cellcolor{lightblue}\gbox{\bf 0.561} & \cellcolor{lightblue}\gbox{\bf 0.593} &
\cellcolor{lightblue}\gbox{\bf 0.100} & \cellcolor{lightblue}\gbox{\bf 0.042} &
\cellcolor{lightblue}\gbox{\bf 0.031} & \cellcolor{lightblue}\gbox{\bf 0.005} &
\cellcolor{lightblue}0.617 & \cellcolor{lightblue}\gbox{\bf 0.692} \\
& \xmark & \textit{500} &
0.042 & 0.021 & 0.024 & \bf 0.004 & 0.556 & 0.585 &
0.121 & 0.054 & 0.050 & \bf 0.006 & 0.613 & 0.684 \\
& \cellcolor{lightblue}\cmark & \cellcolor{lightblue}{} &
\cellcolor{lightblue}\gbox{\bf 0.033} & \cellcolor{lightblue}\gbox{\bf 0.016} &
\cellcolor{lightblue}\gbox{\bf 0.023} & \cellcolor{lightblue}\gbox{\bf 0.004} &
\cellcolor{lightblue}\gbox{\bf 0.558} & \cellcolor{lightblue}\gbox{\bf 0.587} &
\cellcolor{lightblue}\gbox{\bf 0.114} & \cellcolor{lightblue}\gbox{\bf 0.050} &
\cellcolor{lightblue}\gbox{\bf 0.040} & \cellcolor{lightblue}0.007 &
\cellcolor{lightblue}\gbox{\bf 0.615} & \cellcolor{lightblue}\gbox{\bf 0.687} \\

    \bottomrule[0.17em]
\end{tabular}
}
\label{tab:mv_recon_long}
\end{table}

\subsection{Camera Pose Estimation}
\label{sec:exp_pose}

\begin{figure}[t]
  \centering
  \includegraphics[width=\textwidth]{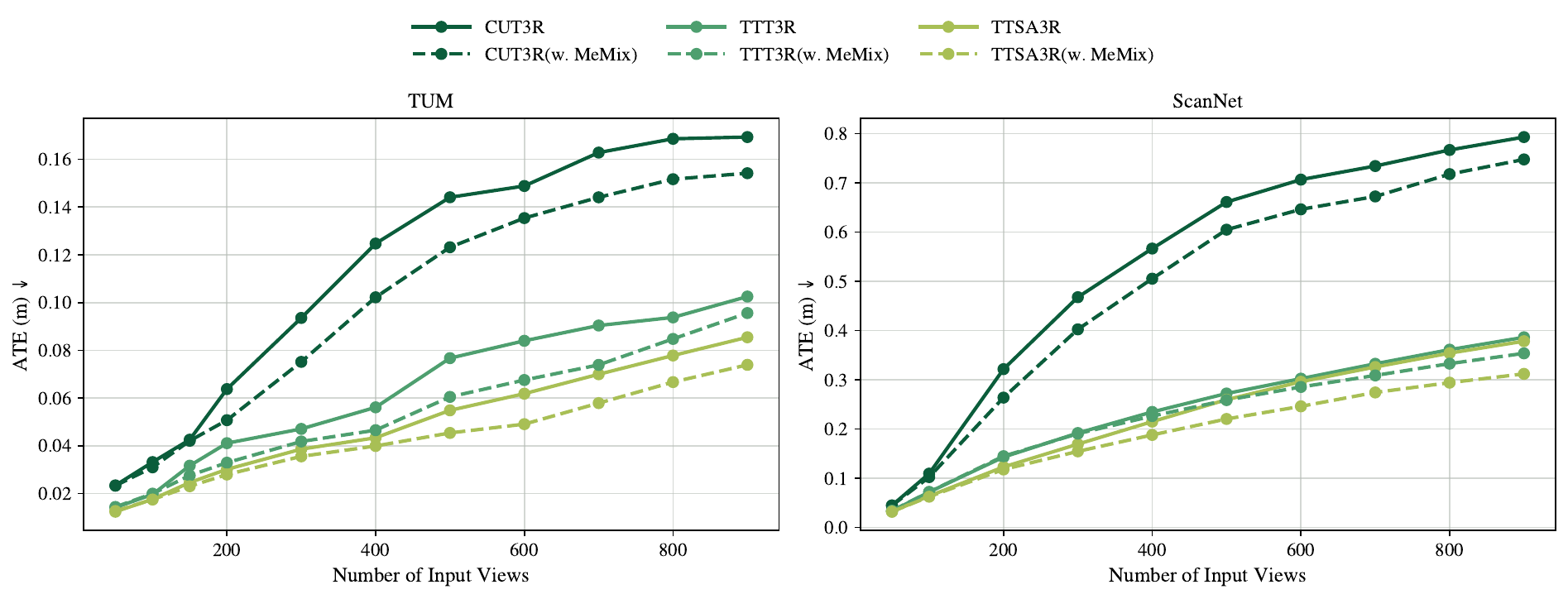}
  \caption{\textbf{Long-sequence pose estimation on TUM and ScanNet.}
  We compare CUT3R, TTT3R, and TTSA3R with their MeMix variants on long input streams, and report the absolute trajectory error (ATE) as the number of input views increases.}
  \label{fig:video_pose}
\end{figure}

Following prior works\cite{wang2025cut3r,chen2026ttt3r,zheng2026ttsa3r,shen2025mut3r}, we further evaluate long-sequence camera pose estimation on both TUM and ScanNet for three recurrent backbones, reporting the absolute trajectory error (ATE) as the number of input views increases; lower is better. The results are summarized in Fig.~\ref{fig:video_pose}.
Across both datasets, MeMix consistently improves pose estimation over the corresponding baselines.

Pose drift in streaming reconstruction is tightly coupled with state degradation. Once the latent state is updated, errors accumulate during readouts. By preserving \texttt{Bottom-k} tokens, MeMix reduces accumulated drift, improving both global trajectory accuracy and frame-to-frame consistency. More results are shown in Table ~\ref{tab:cam_pose} of the supplementary material.

\subsection{Depth Estimation}
\label{sec:exp_depth}

Following common practice\cite{wang2025cut3r,chen2026ttt3r,shen2025mut3r}, we evaluate video depth estimation on KITTI~\cite{datasetkitti}, Bonn~\cite{datasetbonn}, and Sintel~\cite{datasetsintel}. For fair comparison, we keep the same checkpoints and inference settings as their corresponding baselines, and report both scale-invariant and metric-scale metrics in the main paper and supplementary material.

\begin{figure}[t]
  \centering
  \includegraphics[width=\textwidth]{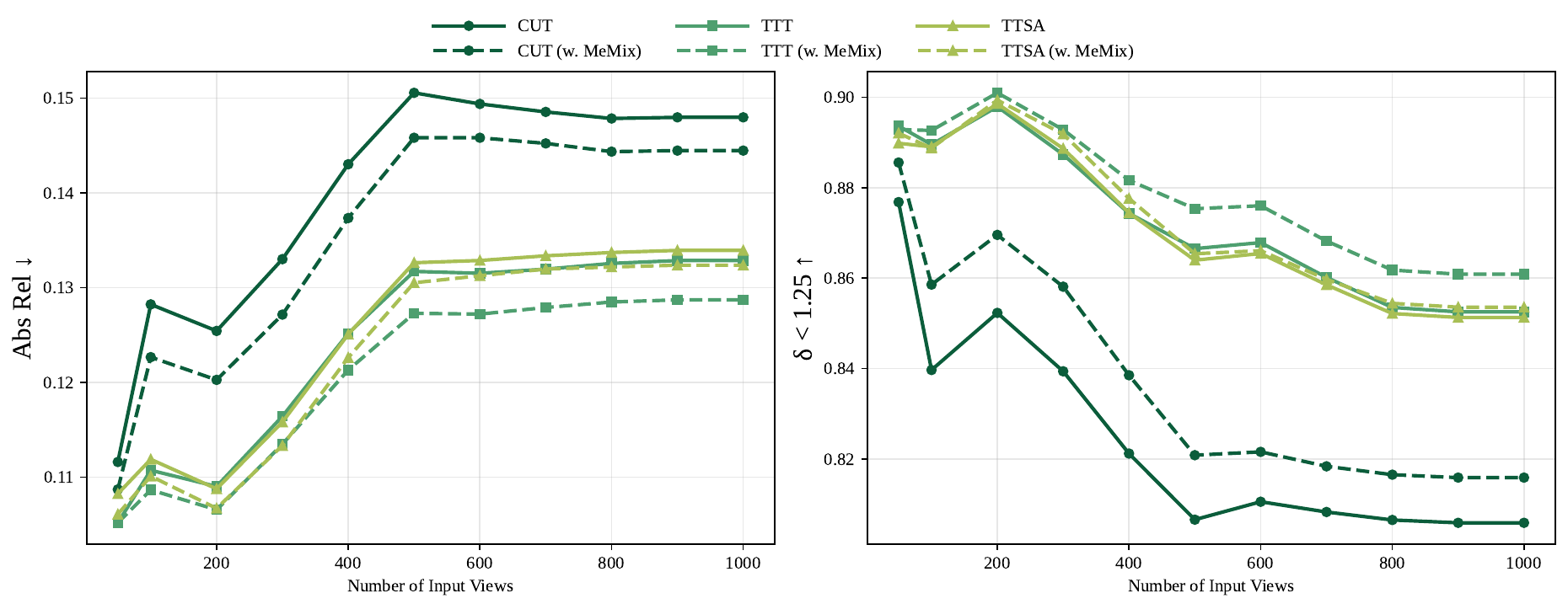}
  \caption{\textbf{Evaluation on Video Depth Estimation.} We compare CUT3R, TTT3R and TTSA3R with their w/o MeMix version under long input streams. MeMix generally improves depth estimation quality for input lengths ranging from 50 to 1000 frames. Notably, the outcomes largely depend on the capacity of the original model.}
  \label{fig:video_depth}
\end{figure}

As shown in Fig.~\ref{fig:video_depth}, MeMix generally improves depth estimation as the input horizon increases. The gains become more evident on longer streams, where recurrent state degradation accumulates over time. At the same time, MeMix also preserves, and in several cases slightly improves, short-horizon performance, suggesting that its benefit is not limited to long-range stability but also comes from more effective state updates. The magnitude of the improvement depends on the strength of the underlying backbone: stronger baselines already mitigate part of the drift, leaving less room for improvement, whereas more fragile baselines benefit more from selective memory writes. More results are shown in Table ~\ref{table:3} of the supplementary material.

\subsection{Ablation Studies}
\label{sec:ablation}

We conduct ablations to isolate how routing (where to write) and write-back (how to write)
affect stability under a fixed-capacity state. Following the analysis protocol used in training-free state
interventions \cite{chen2026ttt3r,zheng2026ttsa3r,shen2025mut3r}, we keep the backbone, patch partition,
and k fixed unless otherwise specified, and report depth,
pose and reconstruction.

\begin{figure}[H]
  \centering
  \includegraphics[width=\textwidth]{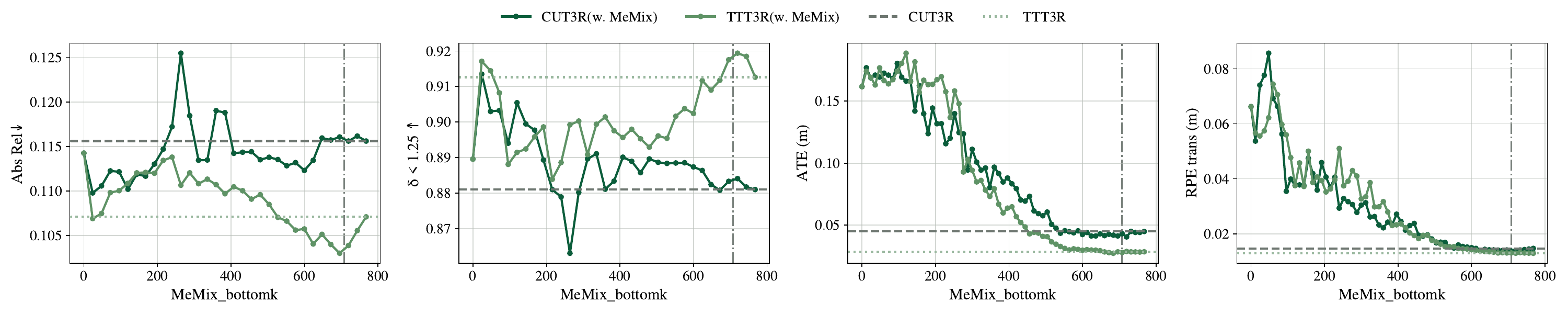}
  \caption{\textbf{Sensitivity to k.}
  We present a representative sweep of k on KITTI video depth estimation and TUM camera pose estimation for CUT3R and TTT3R with/without MeMix.}
  \label{fig:Bottom-k}
\end{figure}

\paragraph{Default configuration and k selection.}
Unless stated otherwise, MeMix uses \texttt{Bottom-k} patch routing with the dot-product score
$S^{\text{dot}}_{t}=\langle \hat{\mathbf{S}}_{t},{\mathbf{Y}}_t\rangle$,
and applies the single-update write-back after the decoder
(optionally gated by $\boldsymbol{\beta}_t$ when enabled by the backbone).
We set k = 708 as a single global default since it is consistently competitive across the reported tasks/metrics;
Fig.~\ref{fig:Bottom-k} presents a representative sweep of k on KITTI\cite{datasetkitti} video depth estimation and TUM camera pose estimation.
Table~\ref{tab:ablation_routing_score_compact} reports this configuration as the default row, and each subsequent block varies one factor while keeping the others fixed.

\begin{table}[t]
\caption{\textbf{Ablations on routing policy and score design.} The Default row corresponds to \texttt{Bottom-k} + \texttt{Dot} + $\text{score}(\hat{\mathbf{S}}_t,{\mathbf{X}}_t)$; other rows change one component at a time.}
\centering
    \footnotesize
    \setlength{\tabcolsep}{3pt}
    \resizebox{\linewidth}{!}{%

\begin{tabular}{@{}l cc ccc ccc}
\toprule
\multirow{2}[2]{*}{\textbf{Variant}} & \multicolumn{2}{c}{\textbf{Depth@KITTI}} & \multicolumn{3}{c}{\textbf{Camera@TUM}} & \multicolumn{3}{c}{\textbf{3R@NRGBD}} \\
\cmidrule(lr){2-3}\cmidrule(lr){4-6}\cmidrule(lr){7-9}
& AbsRel~$\downarrow$ & $\delta<1.25$~$\uparrow$ & ATE~$\downarrow$ & RPE$_\text{trans}$~$\downarrow$ & RPE$_\text{rot}$~$\downarrow$ & Acc~$\downarrow$ & Comp~$\downarrow$ & NC~$\uparrow$ \\
\midrule
Default (TTT3R with MeMix)    & \cellcolor[rgb]{1,1,1}{0.103} & \cellcolor[rgb]{1,1,1}{\textbf{92.1}} & \cellcolor[rgb]{1,1,1}{\textbf{0.028}} & \cellcolor[rgb]{1,1,1}{\textbf{\textbf{0.013}}} & \cellcolor[rgb]{1,1,1}{0.376} & \cellcolor[rgb]{1,1,1}{\textbf{0.099}} & \cellcolor[rgb]{1,1,1}{\textbf{0.020}} & \cellcolor[rgb]{1,1,1}{\textbf{0.616}} \\
\midrule
\multicolumn{9}{@{}l}{\textbf{Patch selection}}\\
Top-k    & \cellcolor[rgb]{1,1,1}{\textbf{0.102}} & \cellcolor[rgb]{1,1,1}{91.8} & \cellcolor[rgb]{1,1,1}{0.068} & \cellcolor[rgb]{1,1,1}{0.021} & \cellcolor[rgb]{1,1,1}{0.595} & \cellcolor[rgb]{1,1,1}{0.178} & \cellcolor[rgb]{1,1,1}{0.049} & \cellcolor[rgb]{1,1,1}{0.587} \\
Random-k & \cellcolor[rgb]{1,1,1}{0.108} & \cellcolor[rgb]{1,1,1}{91.4} & \cellcolor[rgb]{1,1,1}{\textbf{0.028}} & \cellcolor[rgb]{1,1,1}{\textbf{0.013}} & \cellcolor[rgb]{1,1,1}{0.382} & \cellcolor[rgb]{1,1,1}{0.102} & \cellcolor[rgb]{1,1,1}{0.023} & \cellcolor[rgb]{1,1,1}{\textbf{0.616}} \\

\addlinespace[1pt]
\midrule
\multicolumn{9}{@{}l}{\textbf{Scoring function}}\\
Cosine ($S_t^{\text{cos}}$)      & \cellcolor[rgb]{1,1,1}{0.105} & \cellcolor[rgb]{1,1,1}{91.6} & \cellcolor[rgb]{1,1,1}{\textbf{0.028}} & \cellcolor[rgb]{1,1,1}{\textbf{0.013}} & \cellcolor[rgb]{1,1,1}{\textbf{0.375}} & \cellcolor[rgb]{1,1,1}{\textbf{0.099}} & \cellcolor[rgb]{1,1,1}{0.023} & \cellcolor[rgb]{1,1,1}{\textbf{0.616}} \\

Attn \ \ ($S_t^{\text{attn}}$)   & \cellcolor[rgb]{1,1,1}{0.107} & \cellcolor[rgb]{1,1,1}{90.7} & \cellcolor[rgb]{1,1,1}{0.030} & \cellcolor[rgb]{1,1,1}{0.014} & \cellcolor[rgb]{1,1,1}{0.418} & \cellcolor[rgb]{1,1,1}{0.105} & \cellcolor[rgb]{1,1,1}{0.025} & \cellcolor[rgb]{1,1,1}{0.611} \\
\midrule
\multicolumn{9}{@{}l}{\textbf{Update strategy}}\\

Full-update    & \cellcolor[rgb]{1,1,1}{0.108} & \cellcolor[rgb]{1,1,1}{91.4} & \cellcolor[rgb]{1,1,1}{0.035} & \cellcolor[rgb]{1,1,1}{0.015} & \cellcolor[rgb]{1,1,1}{0.443} & \cellcolor[rgb]{1,1,1}{0.127} & \cellcolor[rgb]{1,1,1}{0.034} & \cellcolor[rgb]{1,1,1}{0.604} \\
No-update      & \cellcolor[rgb]{1,1,1}{0.114} & \cellcolor[rgb]{1,1,1}{89.0} & \cellcolor[rgb]{1,1,1}{0.162} & \cellcolor[rgb]{1,1,1}{0.066} & \cellcolor[rgb]{1,1,1}{1.624} & \cellcolor[rgb]{1,1,1}{0.461} & \cellcolor[rgb]{1,1,1}{0.672} & \cellcolor[rgb]{1,1,1}{0.528} \\
\addlinespace[1pt]
\midrule
\multicolumn{9}{@{}l}{\textbf{Routing score}}\\
$\text{score}(\mathbf{S}_{t-1},\mathbf{X}_t)$    & \cellcolor[rgb]{1,1,1}{0.107} & \cellcolor[rgb]{1,1,1}{91.1} & \cellcolor[rgb]{1,1,1}{0.039} & \cellcolor[rgb]{1,1,1}{0.016} & \cellcolor[rgb]{1,1,1}{0.437} & \cellcolor[rgb]{1,1,1}{0.111} & \cellcolor[rgb]{1,1,1}{0.027} & \cellcolor[rgb]{1,1,1}{0.608} \\
$\text{score}(\mathbf{S}_{t-1},\mathbf{Y}_t)$    & \cellcolor[rgb]{1,1,1}{0.124} & \cellcolor[rgb]{1,1,1}{85.9} & \cellcolor[rgb]{1,1,1}{0.041} & \cellcolor[rgb]{1,1,1}{0.016} & \cellcolor[rgb]{1,1,1}{0.426} & \cellcolor[rgb]{1,1,1}{0.249} & \cellcolor[rgb]{1,1,1}{0.055} & \cellcolor[rgb]{1,1,1}{0.575} \\

$\text{score}(\hat{\mathbf{S}}_t,\mathbf{Y}_t)$  & \cellcolor[rgb]{1,1,1}{0.128} & \cellcolor[rgb]{1,1,1}{84.7} & \cellcolor[rgb]{1,1,1}{0.046} & \cellcolor[rgb]{1,1,1}{0.046} & \cellcolor[rgb]{1,1,1}{0.465} & \cellcolor[rgb]{1,1,1}{0.308} & \cellcolor[rgb]{1,1,1}{0.112} & \cellcolor[rgb]{1,1,1}{0.573} \\
\bottomrule
\end{tabular}
}

\label{tab:ablation_routing_score_compact}

\end{table}

\subsubsection*{Patch selection strategy.}
We compare three routing policies: \texttt{Top-k} (update k highest-score patches),
\texttt{Bottom-k} (update k lowest-score patches; Default) and \texttt{Random-k}
(uniformly sample k patches). For patch-level routing, token scores are averaged within each patch prior to selection.

\subsubsection*{Scoring function.}
We evaluate cosine similarity, dot product (default), and attention-derived scores as routing signals:
\begin{equation}
S^{\text{cos}}_{t}=\left\langle \frac{\hat{\mathbf{S}}_{t}}{\|\hat{\mathbf{S}}_{t}\|_2},
\frac{{\mathbf{Y}}_t}{\|{\mathbf{Y}}_t\|_2}\right\rangle,\quad
S^{\text{dot}}_{t}=\langle \hat{\mathbf{S}}_{t},{\mathbf{Y}}_t\rangle
\end{equation}
where $\hat{\mathbf{S}}_{t}$ is the candidate state token for the final decoder and $\mathbf{X}_{t}$
are the image tokens used for routing. For TTT3R-style attention routing \cite{chen2026ttt3r}, we use aggregated
decoder cross-attention as TTT3R\cite{chen2026ttt3r} did.

\subsubsection*{Write-back strategy.}
We compare single-update (one update after the decoder):
\begin{equation}
\mathbf{S}_t= \mathbf{M}_t \odot \hat{\mathbf{S}}_t+(1-\mathbf{M}_t) \odot \mathbf{S}_{t-1}
\end{equation}
and full-update (per-block update inside each decoder layer):
\begin{equation}
\mathbf{S}^{(\ell+1)}_{t}=\mathbf{M}^{(\ell)}_{t} \odot \hat{\mathbf{S}}^{(\ell+1)}_{t} + (1-\mathbf{M}^{(\ell)}_{t}) \mathbf{S}^{(\ell)}_{t}
\end{equation}
together with a no-update (freeze) control.

\subsubsection*{Feature source for routing.}
Let $\mathbf{S}_{t-1}$ be the pre-update state, $\hat{\mathbf{S}}_t$ the decoder-final candidate state,
$\mathbf{X}_t$ the raw image tokens, and $\mathbf{Y}_t$ the interaction tokens used for routing.
We further compare $\text{score}(\mathbf{S}_{t-1},\mathbf{X}_t)$, $\text{score}(\hat{\mathbf{S}}_t,\mathbf{Y}_t)$,
$\text{score}(\hat{\mathbf{S}}_t,\mathbf{X}_t)$, and $\text{score}(\mathbf{S}_{t-1},\mathbf{Y}_t)$ to determine which
routing score strategy is the best.

\subsubsection*{Bottom-k Selection.}

In the ablation study, we also evaluate our \texttt{Bottom-k} strategy. We integrate MeMix into CUT3R and TTT3R and compare the results with the original versions to determine the optimal value of k. By sweeping k in 12 token steps, ranging from 0 to 768, the experiments show that MeMix achieves the best performance when k is set to 708.

\subsubsection*{Inference Efficiency}
\noindent
\begin{minipage}[t]{0.5\linewidth}
To enhance the practical applicability of our approach, we conduct an ablation study comparing the inference speed and GPU memory consumption of three methods with and without MeMix. We test on KITTI using whole-frame input and calculate the average results for each scene. As shown in Table~\ref{table:5}, introducing MeMix has negligible impact on both the inference FPS and peak GPU memory usage across all methods. These results demonstrate that MeMix does not introduce additional side effects.
\end{minipage}
\hfill
\begin{minipage}[t]{0.45\linewidth}
  \vspace{-0.8em}
  \centering
  \captionof{table}{\textbf{Efficiency.} Inference FPS and peak GPU memory with/without MeMix.}
  \label{table:5}
  \footnotesize
  \tabcolsep=0.5em
  \resizebox{\linewidth}{!}{%
  \begin{tabular}{@{}c cc cc@{}}
    \toprule
    \textbf{Method} & \multicolumn{2}{c}{\textbf{FPS (f/s)}} & \multicolumn{2}{c}{\textbf{GPU (GB)}} \\
    \cmidrule(lr){1-1}\cmidrule(lr){2-3}\cmidrule(lr){4-5}
    \textbf{w. MeMix} & \xmark & \cmark & \xmark & \cmark \\
    \midrule
    \textbf{\texttt{CUT3R}}  & 14.39 & 14.13 & 5.31 & 5.31 \\
    \textbf{\texttt{TTT3R}}  & 12.72 & 12.81 & 6.96 & 6.96 \\
    \textbf{\texttt{TTSA3R}} & 12.58 & 12.78 & 6.63 & 6.63 \\
    \bottomrule
  \end{tabular}}
\end{minipage}

\section{Conclusion}
\textbf{Summary.} Fully rewriting the recurrent state at each step causes cumulative interference and catastrophic forgetting in long-horizon inference. We identify this fundamental bottleneck in fixed-state streaming 3D reconstruction and propose MeMix: a training-free, plug-in memory update module that recasts the recurrent state as a mixture of memory patches, substantially improving long-sequence reconstruction quality. MeMix seamlessly integrates into mainstream recurrent reconstruction models, consistently improving performance while preserving short-sequence accuracy, with negligible overhead in GPU memory and inference latency.\\
\textbf{Limitations.} Although MeMix surpasses previous main-stream methods in long-horizon inference, we have not tested what happens when the input contains thousands of frames. Inference on kilometer scale is vital for navigation and perception; some methods~\cite{yuan2026infiniteVGGT,cheng2026longstream,zhang2026loger} have achieved this goal, but it still needs more exploration. Moreover, the \texttt{Bottom-k} selection is heuristic, and we have not analyzed the interpretability of this parameter for the state update. In the future, update strategies based on geometric properties or physical scenarios may further enhance the potential of this process.\\
\section*{Acknowledgement}
We thank Xingyu Chen~\cite{chen2026ttt3r}, Wendi Hu, Haonan Zhou, Chengyi Gao for valuable insights and supports during the project. A small step for your devotion, a huge leap for your successors.

\clearpage

\bibliographystyle{unsrtnat}
\bibliography{main}

\clearpage

\section*{Supplementary Material}

\subsubsection*{A1. Comparison between Top-k \& Bottom-k}
\label{appendix:1}
\begin{figure}[H]
  \centering
\includegraphics[width=\textwidth]{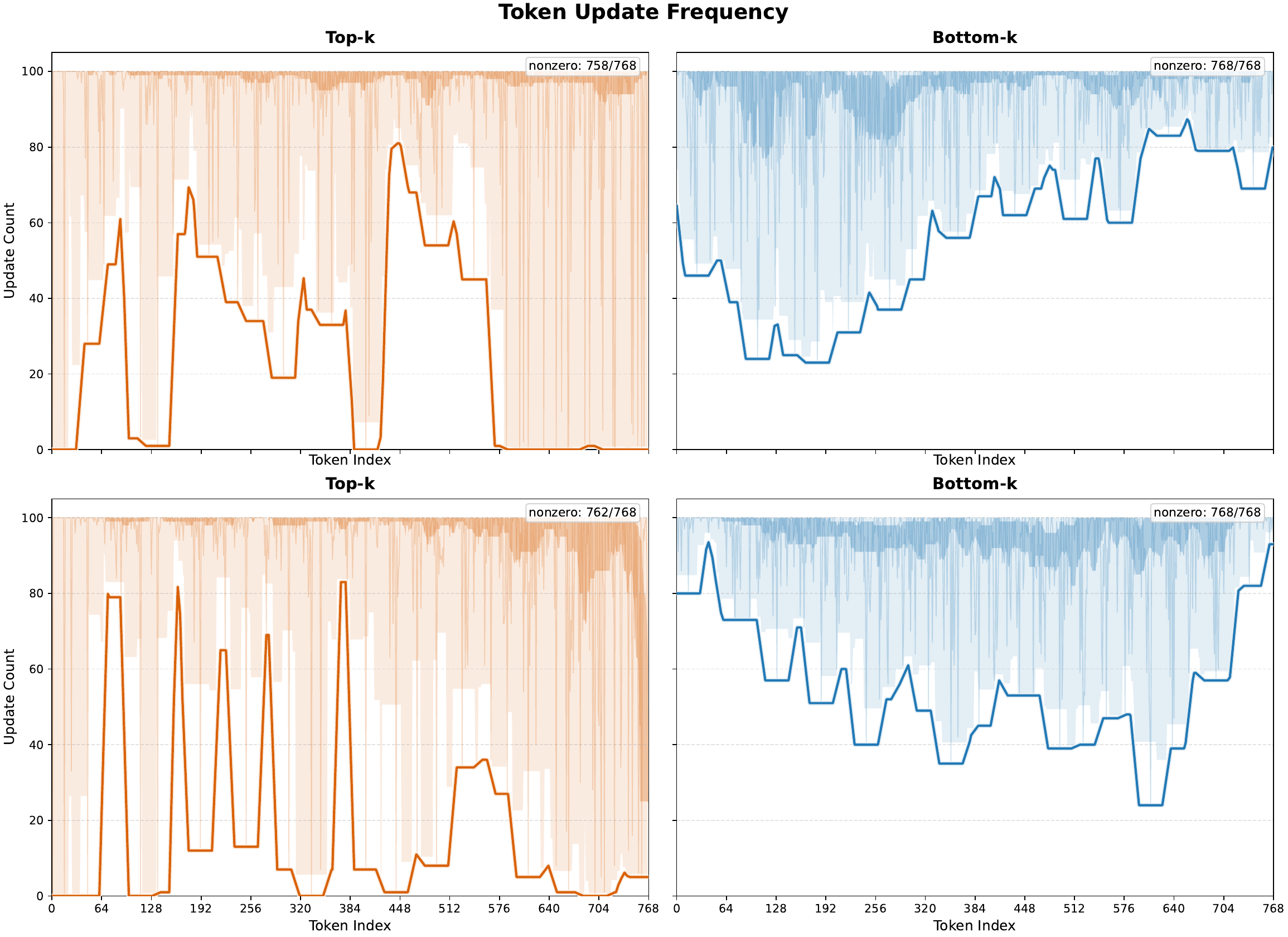}
  \caption{\textbf{Visualization of distinct strategies.} We employ \texttt{Top-k} and \texttt{Bottom-k} strategies separately, and tally the state tokens that are updated in each input frame. Representative examples show that Bottom-k not only achieves a higher update frequency, but also yields a more balanced update distribution across all state tokens.}
  
  \label{S-fig:Bottom-k}
\end{figure}

\texttt{Top-k} updates the most-aligned tokens, creating a positive feedback loop that a small set of high-score tokens is repeatedly selected and reinforced, while the rest receive few updates and gradually become stale. This behavior reduces the effectiveness of memory diversity and utilization. In contrast, \texttt{Bottom-k} updates the least-aligned tokens, which naturally spreads writes across the state over time and improves overall memory coverage. For CUT3R(w. MeMix) on 7-Scenes, \texttt{Top-k} leaves a distinct subset of tokens nearly unupdated, whereas Bottom-k yields far more uniform token updates overall.

\subsubsection*{A2. 3D Reconstruction}

\noindent
We additionally report dense long-sequence 3D reconstruction results on 7-Scenes and NRGBD, using input streams of 300, 400, and 500 frames.
We evaluate reconstruction quality with accuracy, completeness, and normal consistency, where lower accuracy/completeness and higher normal consistency indicate better performance.
Green cells indicate metrics that are improved or preserved relative to the corresponding base model under the same input length.

\noindent
As shown in Table~\ref{S-tab:mv_recon_long}, MeMix consistently improves or preserves dense long-sequence 3D reconstruction performance across different recurrent backbones and datasets, with especially clear gains on CUT3R. These results suggest that MeMix serves as a general memory-update improvement rather than a backbone-specific design.

\begin{table}[h]
\centering
\caption{
    \textbf{3D Reconstruction Results on 7-Scenes~\cite{dataset7scenes} and NRGBD~\cite{datasetnrgbd}.}
     We test MeMix on 7-Scenes and NRGBD, with every frame sampled (Dense Sampling, \textbf{-D)}. \protect\gbox{Green boxes} indicate improved or unchanged performance over the base model (w/o MeMix) under the same input length.
}
\resizebox{\textwidth}{!}{
\begin{tabular}{lcccccccccccccc}
    \toprule[0.17em]
    {\multirow{4}{*}{\textbf{Model}}} &
    {\multirow{4}{*}{\textbf{MeMix}}} &
    {\multirow{4}{*}{\textbf{Input}}} &
    \multicolumn{6}{c}{\textbf{7-Scenes-D}} &
    \multicolumn{6}{c}{\textbf{NRGBD-D}} \\
    \cmidrule(r){4-9} \cmidrule(r){10-15}
    & & &
    \multicolumn{2}{c}{Acc. $\downarrow$} &
    \multicolumn{2}{c}{Comp. $\downarrow$} &
    \multicolumn{2}{c}{NC $\uparrow$} &
    \multicolumn{2}{c}{Acc. $\downarrow$} &
    \multicolumn{2}{c}{Comp. $\downarrow$} &
    \multicolumn{2}{c}{NC $\uparrow$} \\
    \cmidrule(r){4-5} \cmidrule(r){6-7} \cmidrule(r){8-9}
    \cmidrule(r){10-11} \cmidrule(r){12-13} \cmidrule(r){14-15}
    & & &
    Mean & Med. &
    Mean & Med. &
    Mean & Med. &
    Mean & Med. &
    Mean & Med. &
    Mean & Med. \\
    \midrule[0.08em]

\multirow{3}{*}{VGGT \textit{(Offline)}~\cite{wang2025vggt}} & -- & \textit{300} &
\textcolor{lightgray}{\textit{OOM}} & \textcolor{lightgray}{\textit{OOM}} &
\textcolor{lightgray}{\textit{OOM}} & \textcolor{lightgray}{\textit{OOM}} &
\textcolor{lightgray}{\textit{OOM}} & \textcolor{lightgray}{\textit{OOM}} &
\textcolor{lightgray}{\textit{OOM}} & \textcolor{lightgray}{\textit{OOM}} &
\textcolor{lightgray}{\textit{OOM}} & \textcolor{lightgray}{\textit{OOM}} &
\textcolor{lightgray}{\textit{OOM}} & \textcolor{lightgray}{\textit{OOM}} \\
& -- & \textit{400} &
\textcolor{lightgray}{\textit{OOM}} & \textcolor{lightgray}{\textit{OOM}} &
\textcolor{lightgray}{\textit{OOM}} & \textcolor{lightgray}{\textit{OOM}} &
\textcolor{lightgray}{\textit{OOM}} & \textcolor{lightgray}{\textit{OOM}} &
\textcolor{lightgray}{\textit{OOM}} & \textcolor{lightgray}{\textit{OOM}} &
\textcolor{lightgray}{\textit{OOM}} & \textcolor{lightgray}{\textit{OOM}} &
\textcolor{lightgray}{\textit{OOM}} & \textcolor{lightgray}{\textit{OOM}} \\
& -- & \textit{500} &
\textcolor{lightgray}{\textit{OOM}} & \textcolor{lightgray}{\textit{OOM}} &
\textcolor{lightgray}{\textit{OOM}} & \textcolor{lightgray}{\textit{OOM}} &
\textcolor{lightgray}{\textit{OOM}} & \textcolor{lightgray}{\textit{OOM}} &
\textcolor{lightgray}{\textit{OOM}} & \textcolor{lightgray}{\textit{OOM}} &
\textcolor{lightgray}{\textit{OOM}} & \textcolor{lightgray}{\textit{OOM}} &
\textcolor{lightgray}{\textit{OOM}} & \textcolor{lightgray}{\textit{OOM}} \\
\midrule[0.08em]

\multirow{3}{*}{StreamVGGT~\cite{zhuo2026streamVGGT}} & -- & \textit{300} &
\textcolor{lightgray}{\textit{OOM}} & \textcolor{lightgray}{\textit{OOM}} &
\textcolor{lightgray}{\textit{OOM}} & \textcolor{lightgray}{\textit{OOM}} &
\textcolor{lightgray}{\textit{OOM}} & \textcolor{lightgray}{\textit{OOM}} &
\textcolor{lightgray}{\textit{OOM}} & \textcolor{lightgray}{\textit{OOM}} &
\textcolor{lightgray}{\textit{OOM}} & \textcolor{lightgray}{\textit{OOM}} &
\textcolor{lightgray}{\textit{OOM}} & \textcolor{lightgray}{\textit{OOM}} \\
& -- & \textit{400} &
\textcolor{lightgray}{\textit{OOM}} & \textcolor{lightgray}{\textit{OOM}} &
\textcolor{lightgray}{\textit{OOM}} & \textcolor{lightgray}{\textit{OOM}} &
\textcolor{lightgray}{\textit{OOM}} & \textcolor{lightgray}{\textit{OOM}} &
\textcolor{lightgray}{\textit{OOM}} & \textcolor{lightgray}{\textit{OOM}} &
\textcolor{lightgray}{\textit{OOM}} & \textcolor{lightgray}{\textit{OOM}} &
\textcolor{lightgray}{\textit{OOM}} & \textcolor{lightgray}{\textit{OOM}} \\
& -- & \textit{500} &
\textcolor{lightgray}{\textit{OOM}} & \textcolor{lightgray}{\textit{OOM}} &
\textcolor{lightgray}{\textit{OOM}} & \textcolor{lightgray}{\textit{OOM}} &
\textcolor{lightgray}{\textit{OOM}} & \textcolor{lightgray}{\textit{OOM}} &
\textcolor{lightgray}{\textit{OOM}} & \textcolor{lightgray}{\textit{OOM}} &
\textcolor{lightgray}{\textit{OOM}} & \textcolor{lightgray}{\textit{OOM}} &
\textcolor{lightgray}{\textit{OOM}} & \textcolor{lightgray}{\textit{OOM}} \\
\midrule[0.08em]

\multirow{6}{*}{CUT3R~\cite{wang2025cut3r}} & \xmark & \textit{300} &
0.099 & 0.062 & 0.048 & 0.014 & 0.542 & 0.562  &
0.137 & 0.092 & 0.066 & 0.024 & 0.572 & 0.609 \\
& \cellcolor{lightblue}\cmark & \cellcolor{lightblue}{} &
\cellcolor{lightblue}\gbox{0.076} & \cellcolor{lightblue}\gbox{0.045} &
\cellcolor{lightblue}\gbox{0.039} & \cellcolor{lightblue}\gbox{0.010} &
\cellcolor{lightblue}\gbox{0.549} & \cellcolor{lightblue}\gbox{0.573} &
\cellcolor{lightblue}\gbox{0.113} & \cellcolor{lightblue}\gbox{0.081} &
\cellcolor{lightblue}\gbox{0.060} & \cellcolor{lightblue}0.035 &
\cellcolor{lightblue}\gbox{0.578} & \cellcolor{lightblue}\gbox{0.618} \\
& \xmark & \textit{400} &
0.150 & 0.093 & 0.090 & 0.037 & 0.531 & 0.543 &
0.225 & 0.155 & 0.119 & 0.076 & 0.554 & 0.579 \\
& \cellcolor{lightblue}\cmark & \cellcolor{lightblue}{} &
\cellcolor{lightblue}\gbox{0.117} & \cellcolor{lightblue}\gbox{0.071} &
\cellcolor{lightblue}\gbox{0.056} & \cellcolor{lightblue}\gbox{0.015} &
\cellcolor{lightblue}\gbox{0.536} & \cellcolor{lightblue}\gbox{0.552} &
\cellcolor{lightblue}\gbox{0.196} & \cellcolor{lightblue}\gbox{0.128} &
\cellcolor{lightblue}\gbox{0.098} & \cellcolor{lightblue}\gbox{0.062} &
\cellcolor{lightblue}\gbox{0.572} & \cellcolor{lightblue}\gbox{0.609} \\
& \xmark & \textit{500} &
0.165 & 0.114 & 0.094 & 0.039 & 0.522 & 0.531 &
0.313 & 0.203 & 0.202 & 0.148 & 0.554 & 0.580 \\
& \cellcolor{lightblue}\cmark & \cellcolor{lightblue}{} &
\cellcolor{lightblue}\gbox{0.146} & \cellcolor{lightblue}\gbox{0.094} &
\cellcolor{lightblue}\gbox{0.067} & \cellcolor{lightblue}\gbox{0.022} &
\cellcolor{lightblue}\gbox{0.528} & \cellcolor{lightblue}\gbox{0.541} &
\cellcolor{lightblue}\gbox{0.273} & \cellcolor{lightblue}\gbox{0.173} &
\cellcolor{lightblue}\gbox{0.162} & \cellcolor{lightblue}\gbox{0.110} &
\cellcolor{lightblue}\gbox{0.568} & \cellcolor{lightblue}\gbox{0.602} \\
\midrule[0.08em]

\multirow{6}{*}{TTT3R~\cite{chen2026ttt3r}} & \xmark & \textit{300} &
0.030 & 0.016 & 0.019 & \bf 0.004 & 0.558 & 0.588 &
0.057 & 0.035 & 0.016 & \bf 0.003 & 0.595 & 0.650 \\
& \cellcolor{lightblue}\cmark & \cellcolor{lightblue}{} &
\cellcolor{lightblue}\gbox{0.030} & \cellcolor{lightblue}\gbox{0.016} &
\cellcolor{lightblue}\gbox{0.019} & \cellcolor{lightblue}\gbox{\bf 0.004} &
\cellcolor{lightblue}\gbox{\bf 0.559} & \cellcolor{lightblue}\gbox{0.589} &
\cellcolor{lightblue}\gbox{0.052} & \cellcolor{lightblue}\gbox{0.032} &
\cellcolor{lightblue}\gbox{0.015} & \cellcolor{lightblue}\gbox{\bf 0.003} &
\cellcolor{lightblue}\gbox{0.599} & \cellcolor{lightblue}\gbox{0.656} \\
& \xmark & \textit{400} &
0.044 & 0.026 & 0.024 & \bf 0.004 & 0.551 & 0.577 &
0.093 & 0.053 & 0.018 & \bf 0.003 & 0.587 & 0.635 \\
& \cellcolor{lightblue}\cmark & \cellcolor{lightblue}{} &
\cellcolor{lightblue}\gbox{0.039} & \cellcolor{lightblue}\gbox{0.023} &
\cellcolor{lightblue}{0.025} & \cellcolor{lightblue}\gbox{\bf 0.004} &
\cellcolor{lightblue}\gbox{0.552} & \cellcolor{lightblue}\gbox{0.578} &
\cellcolor{lightblue}\gbox{0.078} & \cellcolor{lightblue}\gbox{0.042} &
\cellcolor{lightblue}\gbox{0.016} & \cellcolor{lightblue}\gbox{\bf 0.003} &
\cellcolor{lightblue}\gbox{0.592} & \cellcolor{lightblue}\gbox{0.644} \\
& \xmark & \textit{500} &
0.068 & 0.046 & 0.033 & 0.009 & 0.542 & 0.562 &
0.127 & 0.061 & 0.033 & \bf 0.003 & 0.586 & 0.635 \\
& \cellcolor{lightblue}\cmark & \cellcolor{lightblue}{} &
\cellcolor{lightblue}\gbox{0.057} & \cellcolor{lightblue}\gbox{0.039} &
\cellcolor{lightblue}\gbox{0.030} & \cellcolor{lightblue}\gbox{0.008} &
\cellcolor{lightblue}\gbox{0.546} & \cellcolor{lightblue}\gbox{0.568} &
\cellcolor{lightblue}\gbox{0.105} & \cellcolor{lightblue}\gbox{0.048} &
\cellcolor{lightblue}\gbox{0.026} & \cellcolor{lightblue}0.004 &
\cellcolor{lightblue}\gbox{0.586} & \cellcolor{lightblue}0.633 \\
\midrule[0.08em]

\multirow{6}{*}{TTSA3R~\cite{zheng2026ttsa3r}} & \xmark & \textit{300} &
0.023 & 0.011 & 0.018 & \bf 0.004 & 0.558 & 0.588 &
0.039 & \bf 0.022 & 0.011 & \bf 0.003 & \bf 0.606 & \bf 0.669 \\
& \cellcolor{lightblue}\cmark & \cellcolor{lightblue}{} &
\cellcolor{lightblue}\gbox{\bf 0.022} & \cellcolor{lightblue}\gbox{\bf 0.009} &
\cellcolor{lightblue}\gbox{\bf 0.017} & \cellcolor{lightblue}\gbox{\bf 0.004} &
\cellcolor{lightblue}\gbox{\bf 0.559} & \cellcolor{lightblue}\gbox{\bf 0.588} &
\cellcolor{lightblue}\gbox{\bf 0.037} & \cellcolor{lightblue}\gbox{\bf 0.022} &
\cellcolor{lightblue}\gbox{\bf 0.010} & \cellcolor{lightblue}\gbox{\bf 0.003} &
\cellcolor{lightblue}0.605 & \cellcolor{lightblue}0.668 \\
& \xmark & \textit{400} &
0.030 & 0.016 & 0.022 & \bf 0.004 & 0.553 & 0.580 &
0.060 & \bf 0.027 & \bf 0.010 & \bf 0.003 & \bf 0.598 & \bf 0.655 \\
& \cellcolor{lightblue}\cmark & \cellcolor{lightblue}{} &
\cellcolor{lightblue}\gbox{\bf 0.025} & \cellcolor{lightblue}\gbox{\bf 0.012} &
\cellcolor{lightblue}\gbox{\bf 0.021} & \cellcolor{lightblue}\gbox{\bf 0.004} &
\cellcolor{lightblue}\gbox{\bf 0.554} & \cellcolor{lightblue}\gbox{\bf 0.581} &
\cellcolor{lightblue}\gbox{\bf 0.059} & \cellcolor{lightblue}\gbox{\bf 0.027} &
\cellcolor{lightblue}\gbox{\bf 0.010} & \cellcolor{lightblue}\gbox{\bf 0.003} &
\cellcolor{lightblue}0.596 & \cellcolor{lightblue}0.651 \\
& \xmark & \textit{500} &
0.045 & 0.029 & 0.025 & \bf 0.004 & 0.545 & 0.567 &
0.085 & 0.034 & 0.020 & \bf 0.003 & \bf 0.596 & \bf 0.651 \\
& \cellcolor{lightblue}\cmark & \cellcolor{lightblue}{} &
\cellcolor{lightblue}\gbox{\bf 0.035} & \cellcolor{lightblue}\gbox{\bf 0.021} &
\cellcolor{lightblue}\gbox{\bf 0.023} & \cellcolor{lightblue}\gbox{\bf 0.004} &
\cellcolor{lightblue}\gbox{\bf 0.548} & \cellcolor{lightblue}\gbox{\bf 0.571} &
\cellcolor{lightblue}\gbox{\bf 0.081} & \cellcolor{lightblue}\gbox{\bf 0.032} &
\cellcolor{lightblue}\gbox{\bf 0.014} & \cellcolor{lightblue}\gbox{\bf 0.003} &
\cellcolor{lightblue}0.595 & \cellcolor{lightblue}0.649 \\

    \bottomrule[0.17em]
\end{tabular}
}
\label{S-tab:mv_recon_long}
\end{table}

\label{appendix:2}

\subsubsection*{A3. Pose Estimation}


Following prior works\cite{wang2025cut3r,chen2026ttt3r,shen2025mut3r,zheng2026ttsa3r}, we benchmark camera pose estimation on Sintel \cite{datasetsintel}, TUM-dynamics \cite{datasettum}, and ScanNet \cite{datasetscannet}. Following the short-sequence evaluation protocol adopted in these prior works, we use 50-frame inputs on Sintel and 90-frame inputs on TUM-dynamics and ScanNet, and report standard trajectory metrics including ATE, translational RPE, and rotational RPE.

As shown in Table~\ref{tab:cam_pose}, MeMix largely preserves and often improves pose accuracy over the corresponding recurrent baselines even in these relatively short sequences. This trend is observed across different backbones, indicating that the benefit of sparse memory routing is not limited to very long-horizon inference, but can also improve update quality and reduce drift accumulation under shorter input streams.

\vspace{-1em}
\begin{table}[H]
\caption{
\textbf{Evaluation on Short-Sequence Pose Estimation.} To show that MeMix does not undermine performance under short-sequence evaluation, we evaluate on three datasets using input clips shorter than 100 frames. \protect\gbox{Green boxes} indicate improved or unchanged performance over the corresponding base model (w/o MeMix).
}
\centering
\footnotesize
\renewcommand{\arraystretch}{1.}
\renewcommand{\tabcolsep}{2.5pt}
\resizebox{\linewidth}{!}{
\begin{tabular}{@{}clcccc|ccc|ccc@{}}
\toprule
& & & \multicolumn{3}{c}{\textbf{TUM-dynamics (90 frames)}} & \multicolumn{3}{c}{\textbf{ScanNet (90 frames)}} & \multicolumn{3}{c}{\textbf{Sintel (50 frames)}} \\
\cmidrule(lr){4-6} \cmidrule(lr){7-9} \cmidrule(lr){10-12}
{} & {\textbf{Method}}  & \textbf{Online} & {ATE $\downarrow$} & {RPE trans $\downarrow$} & {RPE rot $\downarrow$} & {ATE $\downarrow$} & {RPE trans $\downarrow$} & {RPE rot $\downarrow$} & {ATE $\downarrow$} & {RPE trans $\downarrow$} & {RPE rot $\downarrow$} \\
\midrule

& Robust-CVD~\cite{kopf2021robustCVD} &\xmark & 0.153 & 0.026 & 3.528 & 0.227 & 0.064 & 7.374 & 0.360 & 0.154 & 3.443 \\
 & CasualSAM~\cite{zhang2022casualSAM} &\xmark & 0.071 & 0.010 & 1.712 & 0.158 & 0.034 & 1.618 & 0.141 & 0.035 & 0.615 \\
 & DUSt3R~\cite{wang2024dust3r} &\xmark & 0.083 & 0.017 & 3.567 & 0.081 & 0.028 & 0.784 & 0.417 & 0.250 & 5.796 \\
{ }& MASt3R~\cite{leroy2024mast3r} &\xmark & {{0.038}} & {{0.012}} & {{0.448}} & {{0.078}} & {{0.020}} & { {0.475}} & {{0.185}} & {0.060} & {1.496} \\
& MonST3R~\cite{zhang2025monst3r}  &\xmark & {{0.098}} & {{0.019}} & {0.935} & {0.077} & {0.018} & {{0.529}} &  {{0.111}} & {0.044} & {0.869} \\

& Easi3R~\cite{chen2025easi3r}  &\xmark & {{0.105}} & {{0.022}} & {{1.064}} & {{0.061}} & {{0.017}} & {0.525} &  {{0.110}} & {0.042} & {0.758} \\

& AETHER\cite{zhu2025aether}  &\xmark & {{0.092}} & {{0.012}} & {{1.106}} & {{0.176}} & {{0.028}} & {{1.204}} &  {{0.189}} & {0.054} & {0.694} \\

& VGGT~\cite{wang2025vggt}  &\xmark & {\bf {0.012}} & {\bf {0.010}} & {\bf {0.310}} & {\bf {0.035}} & {\bf {0.015}} & {\bf {0.377}} & {\bf {0.172}} & \bf {0.062} & \bf {0.471} \\

\midrule

& Spann3R~\cite{wang2025spann3r} &\cmark   & {{0.056}} & {{0.021}} & {{0.591}} & {0.096} & {{0.023}} & {{0.661}} & {{0.329}} & {0.110} & {4.471} \\

& Point3R~\cite{wu2025point3r} & \cmark  & {0.075} & 0.029 & 0.642 &{0.106} & {0.035} & {1.946} & 0.351 & 0.128 & 1.822\\

& StreamVGGT~\cite{zhuo2026streamVGGT} & \cmark  & {0.061} & 0.033 & 3.209 &{0.161} & {0.057} & {3.647} & 0.251 & 0.149 & 1.894\\

& CUT3R~\cite{wang2025cut3r} & \cmark  & 0.045 & {0.015} & 0.443 & 0.096 & 0.022 & 0.600 & 0.210 & \bf 0.069 & 0.628 \\

\rowcolor{lightblue}\cellcolor{white} & {CUT3R(w. MeMix)} & \cmark  &
\gbox{0.043} & \gbox{0.014} & \gbox{0.424} &
\gbox{0.090} & \gbox{0.022} & 0.604 &
\gbox{\bf 0.190} & 0.075 & \gbox{\bf 0.627}\\

&  TTT3R\cite{chen2026ttt3r} & \cmark  & 0.029 & \bf 0.013 &  0.380 & 0.065 & \bf 0.021 & 0.640 & 0.208 &  0.093 &  0.725 \\

\rowcolor{lightblue}\cellcolor{white} & {TTT3R(w. MeMix)} & \cmark  &
\gbox{0.028} & \bf \gbox{0.013} & \gbox{0.376} &
\gbox{0.065} & {\bf \gbox{0.021}} & 0.677 &
0.210 & \gbox{0.083} & 0.733\\

&  TTSA3R\cite{zheng2026ttsa3r} & \cmark  & 0.026 & \bf 0.013 & \bf 0.372 & 0.058 & \bf 0.021 & \bf 0.561 & {0.210} &  0.084 &  0.738 \\

\rowcolor{lightblue}\cellcolor{white} & {TTSA3R(w. MeMix)} & \cmark  &
\gbox{\bf 0.025} & \bf \gbox{0.013} & \gbox{\bf 0.372} &
\gbox{\bf 0.057} & {\bf \gbox{0.021}} & 0.569 &
\gbox{0.209} & \gbox{0.084} & 0.763\\

\bottomrule
\end{tabular}
}
\label{tab:cam_pose}
\end{table}

\subsubsection*{A4. Video Depth Estimation}
\label{appendix:4}

Following common practice\cite{wang2025cut3r,chen2026ttt3r,shen2025mut3r,zheng2026ttsa3r}, we evaluate on KITTI \cite{datasetkitti}, Bonn \cite{datasetbonn}, and Sintel \cite{datasetsintel}. We use 110-frame input sequences for KITTI and Bonn, and 50-frame input sequences for Sintel. We keep the same settings of each baseline, and report both scale-invariant and metric-scale metrics.

Table~\ref{table:3} shows that MeMix also brings clear gains under these short-sequence settings. Across CUT3R, TTT3R, and TTSA3R, introducing MeMix largely preserves and often improves the corresponding depth metrics, demonstrating that the advantage of sparse routing is not only from better long-range stability, but also from more effective state updates even over shorter horizons.
\vspace{-1em}
\begin{table}[h]
    \caption{\textbf{Video Depth Estimation.} We evaluate scale-invariant and metric depth accuracy on KITTI \cite{datasetkitti}, Sintel \cite{datasetsintel}, and Bonn \cite{datasetbonn} datasets. Methods that require global alignment are denoted as "GA" and \protect\gbox{green boxes} indicate metrics that improve compared to corresponding base model (w/o MeMix).}
    \label{table:3}
    \centering
    \footnotesize
    \setlength{\tabcolsep}{3pt}
    \resizebox{\linewidth}{!}{
    
        \newcommand{\bcell}[1]{\cellcolor{lightblue}#1}
        \newcommand{\bbar}[1]{\multicolumn{1}{c|}{\cellcolor{lightblue}#1}} 

        \begin{tabular}{ccc|cc|cc|cc}
            \toprule[1pt]
            \multirow{2}{*}{\textbf{Alignment}} &
            \multirow{2}{*}{\textbf{Method}} &
            \multirow{2}{*}{\textbf{Online}} &
            \multicolumn{2}{c|}{\textbf{KITTI (110 frames)}} &
            \multicolumn{2}{c|}{\textbf{Sintel (50 frames)}} &
            \multicolumn{2}{c}{\textbf{Bonn (110 frames)}} \\
            \cmidrule(lr){4-5} \cmidrule(lr){6-7} \cmidrule(lr){8-9}
            & & & Abs Rel $\downarrow$ & $\delta<1.25 \uparrow$ & Abs Rel $\downarrow$ & $\delta<1.25 \uparrow$ & Abs Rel $\downarrow$ & $\delta <1.25 \uparrow$ \\
            \midrule[0.6pt]

            \multirow{15}{*}{\textbf{Per-sequence scale}}
                & DUSt3R-GA \cite{wang2024dust3r} & \xmark & 0.144 & \multicolumn{1}{c|}{81.3} & 0.656 & \multicolumn{1}{c|}{45.2} & 0.155 & 83.3 \\
                & MASt3R-GA \cite{leroy2024mast3r} & \xmark & 0.183 & \multicolumn{1}{c|}{74.5} & 0.641 & \multicolumn{1}{c|}{43.9} & 0.252 & 70.1 \\
                & MonST3R-GA \cite{zhang2025monst3r} & \xmark & 0.168 & \multicolumn{1}{c|}{74.4} & 0.378 & \multicolumn{1}{c|}{55.8} & 0.067 & 96.3 \\
                & Easi3R \cite{chen2025easi3r} & \xmark & 0.102 & \multicolumn{1}{c|}{91.2} & 0.377 & \multicolumn{1}{c|}{55.9} & 0.059 & 97.0 \\
                & VGGT \cite{wang2025vggt} & \xmark &
                    \textbf{0.070} & \multicolumn{1}{c|}{\textbf{96.5}} &
                    \textbf{0.287} & \multicolumn{1}{c|}{\textbf{66.1}} &
                    \textbf{0.055} & \textbf{97.1} \\
                \cmidrule[0.6pt]{2-9}
                & Spann3R \cite{wang2025spann3r} & \cmark & 0.198 & \multicolumn{1}{c|}{73.7} & 0.622 & \multicolumn{1}{c|}{42.6} & 0.144 & 81.3 \\
                & Point3R \cite{wu2025point3r} & \cmark & 0.136 & \multicolumn{1}{c|}{84.2} & 0.452 & \multicolumn{1}{c|}{48.9} & 0.060 & 96.0 \\
                & STREAM3R$^{\alpha}$ \cite{lan2026stream3r} & \cmark & 0.116 & \multicolumn{1}{c|}{89.6} & 0.478 & \multicolumn{1}{c|}{51.1} & 0.075 & 94.1 \\
                & StreamVGGT \cite{zhuo2026streamVGGT} & \cmark & 0.173 & \multicolumn{1}{c|}{72.1} &
                    \textbf{0.323} & \multicolumn{1}{c|}{\textbf{65.7}} &
                    \textbf{0.059} & \textbf{97.2} \\
                & CUT3R \cite{wang2025cut3r} & \cmark & {0.116} & \multicolumn{1}{c|}{88.1} & 0.426 & \multicolumn{1}{c|}{47.3} & 0.079 & 93.7 \\

                & \bcell{CUT3R(w. MeMix)} & \bcell{\cmark} &
                    \bcell{\gbox{0.115}} & \bbar{\gbox{88.6}} &
                    \bcell{0.436} & \bbar{46.2} &
                    \bcell{\gbox{0.078}} & \bcell{\gbox{93.8}} \\

                & TTT3R \cite{chen2026ttt3r} & \cmark & 0.107 & \multicolumn{1}{c|}{91.2} & 0.409 & \multicolumn{1}{c|}{48.9} & 0.069 & 95.5 \\

                & \bcell{TTT3R(w. MeMix)} & \bcell{\cmark} &
                    \bcell{\gbox{\textbf{0.103}}} & \bbar{\gbox{92.1}} &
                    \bcell{\gbox{0.407}} & \bbar{\gbox{49.2}} &
                    \bcell{0.070} & \bcell{95.1} \\

                & TTSA3R \cite{zheng2026ttsa3r} & \cmark & \textbf{0.103} & \multicolumn{1}{c|}{91.9} & 0.410 & \multicolumn{1}{c|}{49.6} & 0.064 & 96.4 \\

                & \bcell{TTSA3R(w. MeMix)} & \bcell{\cmark} &
                    \bcell{\gbox{\textbf{0.103}}} & \bbar{\gbox{\textbf{92.2}}} &
                    \bcell{\gbox{0.400}} & \bbar{\gbox{50.2}} &
                    \bcell{0.065} & \bcell{96.0} \\

            \midrule[0.6pt]

            \multirow{9}{*}{\textbf{Metric scale}}
                & MASt3R-GA \cite{leroy2024mast3r} & \xmark & 0.467 & \multicolumn{1}{c|}{15.2} & 1.022 & \multicolumn{1}{c|}{14.3} & 0.272 & 70.6 \\
                & Point3R \cite{wu2025point3r} & \cmark & 0.191 & \multicolumn{1}{c|}{73.8} & \textbf{0.777} & \multicolumn{1}{c|}{17.1} & 0.137 & 94.7 \\
                & STREAM3R$^{\alpha}$ \cite{lan2026stream3r} & \cmark & 0.234 & \multicolumn{1}{c|}{57.6} & 1.041 & \multicolumn{1}{c|}{21.0} & 0.084 & 94.4 \\
                & CUT3R \cite{wang2025cut3r} & \cmark & 0.129 & \multicolumn{1}{c|}{82.8} & 1.020 & \multicolumn{1}{c|}{23.7} & 0.103 & 88.9 \\

                & \bcell{CUT3R(w. MeMix)} & \bcell{\cmark} &
                    \bcell{\gbox{0.122}} & \bbar{\gbox{85.0}} &
                    \bcell{1.068} & \bbar{\gbox{24.1}} &
                    \bcell{0.104} & \bcell{88.8} \\

                & TTT3R \cite{chen2026ttt3r} & \cmark & 0.107 & \multicolumn{1}{c|}{89.2} & 0.978 & \multicolumn{1}{c|}{23.3} & 0.090 & 94.4 \\

                & \bcell{TTT3R(w. MeMix)} & \bcell{\cmark} &
                    \bcell{\gbox{\textbf{0.103}}} & \bbar{\gbox{\textbf{89.9}}} &
                    \bcell{0.984} & \bbar{\gbox{23.6}} &
                    \bcell{0.094} & \bcell{92.9} \\

                & TTSA3R \cite{zheng2026ttsa3r} & \cmark & 0.110 & \multicolumn{1}{c|}{88.6} & 0.959 & \multicolumn{1}{c|}{24.5} & \textbf{0.080} & \textbf{96.4} \\

                & \bcell{TTSA3R(w. MeMix)} & \bcell{\cmark} &
                    \bcell{\gbox{0.107}} & \bbar{\gbox{89.1}} &
                    \bcell{0.962} & \bbar{\gbox{\textbf{24.9}}} &
                    \bcell{0.083} & \bcell{96.1} \\

            \bottomrule[1pt]
        \end{tabular}
    }
\end{table}


\subsubsection*{A5. Visualization}
\label{appendix:5}

\begin{figure}[H]
  \centering
  \includegraphics[width=\textwidth]{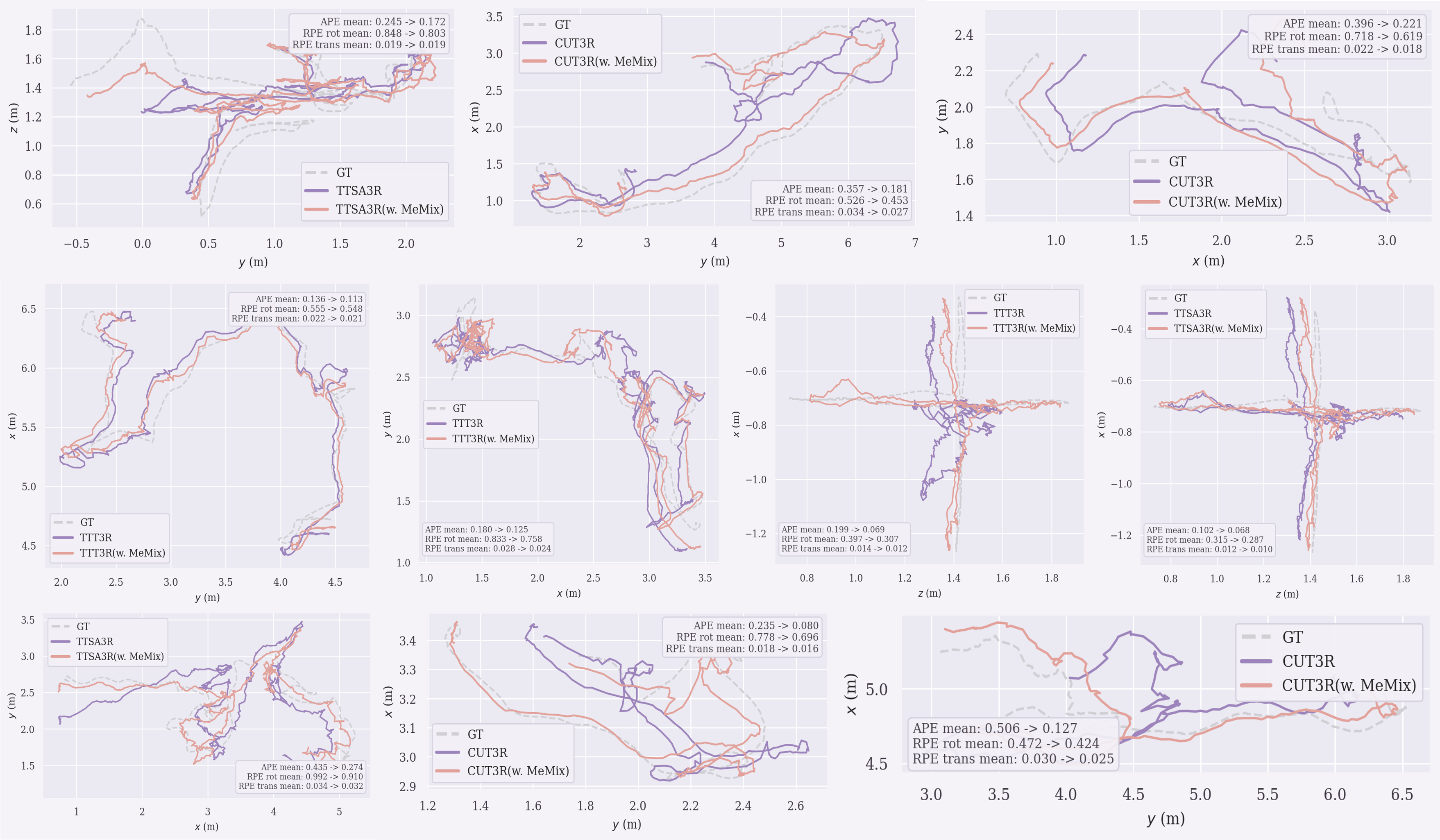}
  \caption{{\bf Visualization of Estimated Camera Trajectories -- Long Sequence.}
We compare \textcolor{gt}{$\bullet$}~\textcolor{gt}{GT},
\textcolor{baseline}{$\bullet$}~\textcolor{baseline}{baseline trajectories},
and \textcolor{memix}{$\bullet$}~\textcolor{memix}{their corresponding MeMix variants}
across three backbones: CUT3R, TTT3R, and TTSA3R.
Across different backbones, the \textcolor{memix}{MeMix} variants generally stay closer to the
\textcolor{gt}{ground-truth trajectories} and exhibit reduced drift over long sequences.}
  \label{fig:visual_pose}
\end{figure}

\noindent
Fig.~\ref{fig:visual_pose} shows qualitative trajectory visualizations on long sequences. Across different backbones, the MeMix variants generally stay closer to the ground-truth trajectories and exhibit reduced drift. More specifically, the improvement is most visible in challenging segments with longer temporal horizons and larger camera motion, where the baseline trajectories tend to gradually deviate from the ground truth or accumulate local drift. In contrast, the MeMix variants better preserve the overall trajectory shape and remain more consistent with the reference path over time. These qualitative observations are consistent with the quantitative ATE improvements reported in Sec.~A3, further supporting that selective memory updates help stabilize long-sequence pose estimation.

\end{document}